\documentclass[10pt,a4paper,logo]{paper}

\usepackage[authoryear,round]{natbib}
\usepackage{dsfont}
\usepackage{tikzducks}
\usepackage{algorithm}
\usepackage{algorithmic}
\usepackage{bm}
\usepackage{xspace}
\usepackage{subcaption}
\usepackage{multirow}
\usepackage{wrapfig}
\usepackage[capitalise,nameinlink]{cleveref}

\definecolor{title_blue}{HTML}{204899}
\definecolor{cite_blue}{HTML}{044dc1}
\definecolor{cite_purple}{HTML}{7406a7}
\hypersetup{
    colorlinks=true,
    citecolor=cite_blue,
    linkcolor=cite_purple,
    urlcolor=cite_purple,
    pdftitle={RACE: Calibrated Cumulative Scoring for Rollout-Free Robot Policy Checkpoint Selection},
    pdfauthor={Changyeon Kim, Yekyung Nah, Kimin Lee}
}

\newlength\savewidth

\definecolor{Blue9}{rgb}{0.098,0.3,0.9}
\definecolor{Red7}{rgb}{0.941, 0.243, 0.243}
\definecolor{Green7}{RGB}{55, 178, 77}
\definecolor{CubsRed}{RGB}{198, 1, 31}
\definecolor{BrickRed}{rgb}{0.6,0,0}
\definecolor{RoyalBlue}{rgb}{0,0,0.8}
\definecolor{BostonGreen}{RGB}{35, 138, 68}
\definecolor{CYRed}{RGB}{228, 0, 43}
\definecolor{CYmagenta}{RGB}{215, 153, 93}
\definecolor{plgray}{HTML}{999999}

\definecolor{baselinecolor}{HTML}{EEEEEE}
\definecolor{algcomment}{gray}{0.45}
\newcommand{\mname}[1]{\textbf{#1}}
\newcommand{\method}{\textsc{RACE}\xspace}
\newcommand{\metfull}{\textit{\underline{R}eference-c\underline{A}librated cumulative scoring for rollout-free \underline{C}heckpoint S\underline{E}lection}\xspace}

\title{\method: Calibrated Cumulative Scoring for Rollout-Free Robot Policy Checkpoint Selection}
\reportnumber{}

\author[1]{Changyeon Kim}
\author[1]{Yekyung Nah}
\author[1,2]{Kimin Lee}
\affil[1]{KAIST}
\affil[2]{Config}
\paperurl{https://changyeon.site/race/}
\correspondingauthor{Correspondence to: \email{changyeon.kim@kaist.ac.kr}.}

\begin{abstract}

Assessing the deployment performance of learned robot policies during training is challenging, as directly measuring task success requires costly real-world trials for each candidate checkpoint. Averaged validation loss offers a rollout-free alternative, but its checkpoint rankings can be inconsistent with deployment performance. We analyze two limitations of this criterion: it penalizes action deviations without considering whether the policy can recover from them, and gives later reference predictions equal weight even when earlier deviations may lead execution away from the reference. We propose \method, a novel rollout-free metric for ranking checkpoints using a fixed set of held-out successful trajectories. \method calibrates action deviations against state-conditional thresholds derived from action differences at similar states in these trajectories. We aggregate these scores through a running product, so that a substantial deviation reduces the contribution of subsequent reference predictions. Averaging the resulting cumulative scores over time and across trajectories yields the checkpoint score. In simulation, \method outperforms baselines across validation sources, while several baselines degrade sharply when the validation set changes. On real robots, its rankings align with deployment success across the evaluated policy backbones and tasks of varying difficulty. Perturbation analyses further show that \method distinguishes deviations with markedly different effects on task success, even when baselines respond similarly.

\end{abstract}

\begin{document}
\maketitle

\section{Introduction}
    Recent advances in robot learning have enabled manipulation policies to perform an increasingly broad range of tasks across different embodiments~\citep{black2024pi_0,black2025pi,bjorck2025gr00t,team2025gemini,team2025gemini1.5,generalist2026gen1}. As these policies are trained and adapted to new settings, it remains difficult to assess how their ability to complete tasks evolves during training. A decreasing training loss does not necessarily identify the checkpoint that will perform best in deployment. Directly measuring task success requires repeated trials on a real robot, and evaluating the many checkpoints produced during training therefore incurs substantial robot time and human effort. This motivates validation criteria that can identify promising checkpoints and focus hardware evaluation on a small set of candidates.

    To reduce hardware evaluation costs, surrogate-based approaches evaluate policies in calibrated simulators~\citep{li2024evaluating,jain2025polaris} or learned world models~\citep{li2025worldeval,quevedo2025worldgym,wang2026interactive, zhang2026real}. However, their usefulness depends on how accurately the surrogate represents the target environment, and constructing or adapting it can require additional data and computation. Rollout-free evaluation offers a complementary approach: a fixed validation set is collected once and reused across checkpoints, whereas deployment trials must be repeated for each evaluated candidate. A common criterion is prediction error on held-out demonstrations, although it can correlate poorly with closed-loop success~\citep{li2024evaluating, pan2025muchado}.

    This mismatch reflects two limitations. First, smaller action errors do not necessarily imply higher task success: a closed-loop policy can correct some deviations, whereas others prevent task completion. Reducing a recoverable deviation can therefore lower validation loss without increasing success. Second, held-out evaluation continues along the reference trajectory regardless of the policy's previous predictions. After a substantial deviation, later reference observations may no longer reflect the states the policy would encounter~\citep{ross2011reduction,spencer2021feedback,simchowitz2025pitfalls}, yet a temporal mean gives their scores equal weight. Off-manifold norm~\citep{pan2025muchado}, a recent alternative, measures deviation from a local expert action manifold but likewise scores predictions pointwise, without accounting for how earlier deviations affect the availability of later execution.

    In this paper, we propose \method (\metfull), a novel rollout-free metric for ranking policy checkpoints highly aligned with deployment performance using a fixed set of held-out successful trajectories. \method calibrates action deviations against state-conditional thresholds derived from action differences at similar states in these trajectories. To compare deviations across action groups with different physical units, such as translation and rotation, we measure each group's deviation separately and normalize it by its threshold. Deviations within the threshold receive a group score of one, while larger relative deviations receive lower scores, avoiding rewards for closer imitation within the chosen range. We combine the group scores into timestep scores and form a cumulative score curve through their running product, so that a substantial deviation reduces the contribution of subsequent reference predictions. Averaging this curve over time preserves the contribution of the trajectory prefix before a deviation while giving earlier reductions a greater effect on the score. Finally, we average the trajectory scores across the reference set, and rank checkpoints in descending order of the resulting score.

    Throughout extensive experiments, we demonstrate the effectiveness of \method in both simulation and real-world settings. In simulation, \method consistently outperforms the baselines across validation sets drawn from demonstrations and policy rollouts, while several baselines deteriorate sharply when the validation source changes. On real robots, its rankings remain positively correlated with deployment success across the evaluated policy backbones and tasks of varying difficulty. Perturbation analyses further show that \method distinguishes perturbations with markedly different effects on task success, even when baselines respond similarly.

    We highlight our key contributions below:
    \begin{itemize}
        \item We identify and analyze two limitations of averaged validation loss for checkpoint selection: it penalizes action deviations without considering whether the policy can recover from them, and temporal averaging ignores how earlier deviations affect the relevance of later reference predictions.
        \item We propose \method, a rollout-free checkpoint scoring method that calibrates action deviations using successful reference trajectories, forms cumulative scores through running products, and averages them over time and across trajectories.
        \item We demonstrate improved checkpoint ranking in standard held-out demonstration evaluation for both simulation and real-world manipulation tasks.
    \end{itemize}
\section{Related Work}
\label{sec:related_work}
    \paragraph{Evaluating robot policies}
    Prior work has sought to reduce the cost of evaluating robot policies on real hardware~\citep{bauer2022real,atreya2025roboarena,zhou2025autoeval}. Sequential statistical procedures~\citep{snyder2025isyour,snyder2026beyond} reduce the number of trials by stopping policy comparisons once sufficient evidence has accumulated, while still relying on rollout outcomes. Simulation-based~\citep{li2024evaluating,jain2025polaris,zhang2026real} and world-model-based~\citep{li2025worldeval,quevedo2025worldgym,li2026dworldeval} evaluators assess candidates through interactions with a surrogate environment. Hybrid real--sim approaches combine paired hardware and simulation evaluations through control variates~\citep{luo2025leveraging} or prediction-powered inference~\citep{badithela2025reliable} to obtain confidence intervals on mean real-world performance with fewer hardware trials. These methods still rely on rollout outcomes from the policies being evaluated. \method ranks checkpoints from their predictions on a fixed set of held-out successful trajectories, requiring no additional hardware or surrogate rollouts during scoring. The same reference set is reused across candidates to support shortlisting before deployment evaluation.

    \paragraph{Rollout-free policy evaluation}
    Off-policy evaluation (OPE) estimates a policy's expected return from logged data, using learned value functions~\citep{le2019batch,paine2020hyperparameter} or importance-sampling corrections~\citep{precup2000eligibility,jiang2016doubly}. These approaches use rewards in the logged data to estimate return. Our setting targets relative checkpoint ranking from successful reference trajectories, without per-step reward annotations or a learned return estimator.

    Reward-free model selection has also been studied for imitation learning~\citep{hussenot2021hyperparameter}. Validation action MSE is a common criterion, yet it can correlate poorly with deployment performance~\citep{li2024evaluating,pan2025muchado}. It penalizes action discrepancies without considering whether the policy can recover from them. Off-manifold norm~\citep{pan2025muchado} measures the component of a predicted action outside a local subspace formed by expert actions at nearby states. Both criteria score predictions at recorded observations without adjusting the contribution of later predictions for earlier deviations. \method derives thresholds from action differences at similar states in successful reference trajectories and converts deviations beyond these thresholds into lower timestep scores. The running product of these scores forms a cumulative score curve, reducing the contribution of later reference predictions after substantial earlier deviations. We rank checkpoints by averaging cumulative scores over time within each trajectory and then across reference trajectories.

\section{Main Method}
    \label{sec:surval}
    In this section, we introduce \method, a novel rollout-free validation metric for policy checkpoint selection highly correlated with deployment performance. We first formulate the checkpoint selection problem (\Cref{sec:problem}), then illustrate the limitations of averaged prediction error through a motivating example (\Cref{sec:method:motivating_example}), and finally present the scoring method (\Cref{sec:method:surval}).

    \subsection{Problem Formulation}
    \label{sec:problem}

    Training a policy $\pi$ typically yields a sequence of $N_{\text{ckpt}}$ checkpoints $\Pi = \{\pi_1, \dots, \pi_{N_{\text{ckpt}}}\}$ saved at successive gradient steps. Each checkpoint's quality is typically measured by its \emph{closed-loop success rate}:
   \begin{equation*}
     J(\pi)=\Pr_{\rho_0,\pi}\!\left[\exists\,t\le T:\ s_t\in\mathcal S_{\mathrm{success}}\right],
   \end{equation*}
   where $\rho_0$ is the initial-state distribution, $T$ is the episode horizon, and $S_{\mathrm{success}}$ indicates the set of successful states. Because $J(\pi)$ can only be estimated through deployment at environments, evaluating every $\pi$ scales poorly with $|\Pi|$ and may be unsafe on real hardware. We therefore study \emph{rollout-free checkpoint selection}: given a held-out set of successful reference trajectories $\mathcal{D}_{\mathrm{val}}=\{\tau^{(j)}\}_{j=1}^{N_{\mathrm{val}}}$, we seek a scoring function $S(\pi;\mathcal{D}_{\mathrm{val}})$ that ranks candidates without executing rollouts, using only each policy's predictions on $\mathcal{D}_{\mathrm{val}}$, such that its induced ranking is highly correlated with the deployment ranking under $J$. The reference trajectories may be demonstrations or previously collected successful policy rollouts.

    \begin{figure}[!t]
    \centering
    \includegraphics[width=\textwidth]{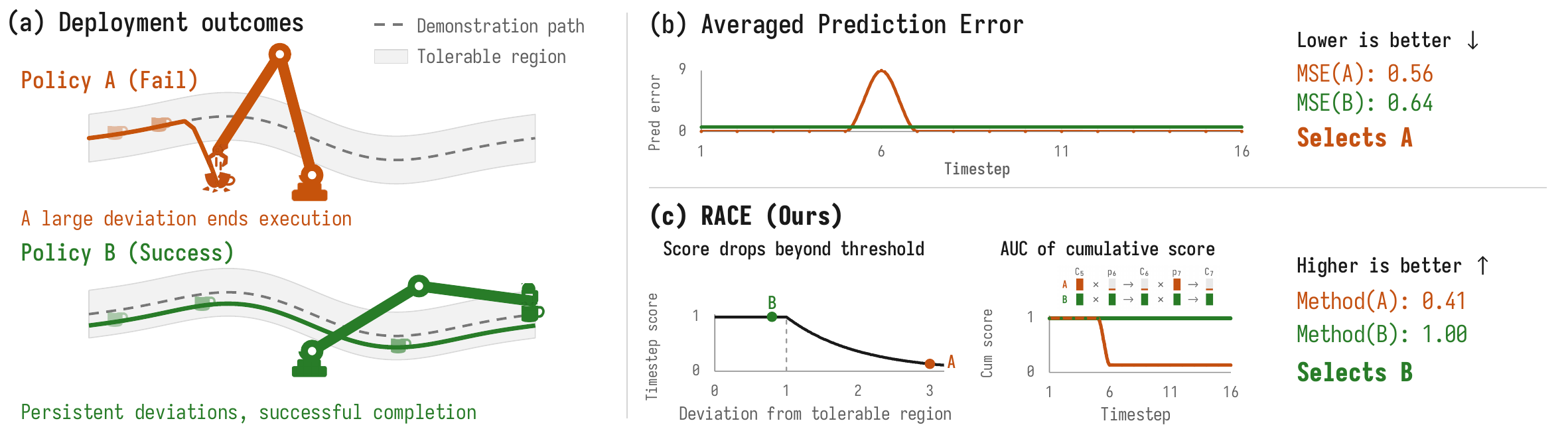}
    \caption{\textbf{Motivation and key idea of \method.} (a) Both policies are evaluated on the held-out demonstration. Policy $\pi_A$ deviates from the demonstrated route and breaks the cup, whereas $\pi_B$ reaches the destination despite persistent smaller deviations. (b) Averaged prediction error dilutes the large deviation of $\pi_A$ with smaller errors elsewhere, giving it a lower mean squared error than $\pi_B$. (c) Deviations within a reference-derived threshold receive a timestep score of one; larger deviations receive lower scores. Running products reduce the contribution of later predictions, as illustrated by two successive updates in the small bars. Averaging the cumulative scores over time correctly selects $\pi_B$; higher scores are preferred.}
    \label{fig:motivating_examples}
\end{figure}

    \subsection{Pitfall of Averaged Prediction Error}
    \label{sec:method:motivating_example}

    A common rollout-free criterion for checkpoint selection is the averaged prediction error on held-out demonstrations~\citep{li2024evaluating, pan2025muchado}. Given a policy $\pi$ and validation set $\mathcal{D}_{\mathrm{val}}$, let $\ell_t = \mathcal{L}(\pi(o_t), a_t)$ denote the per-step prediction error at timestep $t$, where $\mathcal{L}$ is a distance metric matched to the policy's output space (\textit{e.g.}\ MSE for deterministic policies, flow-matching or negative log-likelihood for generative ones). it averages errors within each trajectory and then across trajectories, giving each demonstration equal weight. The criterion selects the checkpoint maximizing:
    \begin{equation*}
      S_{\mathrm{avg}}(\pi;\mathcal D_{\mathrm{val}})
      = -\frac{1}{N_{\mathrm{val}}}
      \sum_{j=1}^{N_{\mathrm{val}}}
      \left[
        \frac{1}{T_j}\sum_{t=1}^{T_j}\ell_t^{(j)}
      \right],
    \end{equation*}
    where $T_j$ denotes the length of trajectory $j$.
    
    \looseness=-1
    To illustrate the limitation, we provide two hypothetical policies evaluated on the same held-out demonstration (\Cref{fig:motivating_examples}). Policy $\pi_A$ closely reproduces the reference actions at most timesteps, but one of its actions causes task failure during execution (\textit{e.g.}, dropping the object beyond the robot’s reach). Policy $\pi_B$ differs more frequently from the reference actions at many timesteps, but it resides in tolerable region and completes the task. Despite these contrasting outcomes, $\pi_A$ can obtain a lower average prediction error because its many small errors dilute the isolated large deviation. The mean-error criterion would then misleadingly select $\pi_A$ over $\pi_B$.

    These observations suggest two considerations for designing rollout-free evaluation criteria. First, action prediction errors need not translate directly into task failure. A closed-loop policy may correct some deviations through subsequent actions, whereas other deviations can prevent task completion. Reducing an error from which the policy can recover therefore need not improve its success rate, even though it lowers the validation loss. This motivates calibrating prediction errors before aggregating them, rather than treating every numerical reduction as an improvement in checkpoint quality. Second, averaged prediction error scores each prediction at a recorded demonstration observation, regardless of the policy’s earlier predictions. If an earlier deviation would cause execution to depart from the reference, predictions at subsequent timesteps may no longer describe the policy’s closed-loop behavior. This motivates reducing the contribution of later validation steps following a substantial deviation.

    \begin{figure}[!t]
    \centering
    \includegraphics[width=\textwidth]{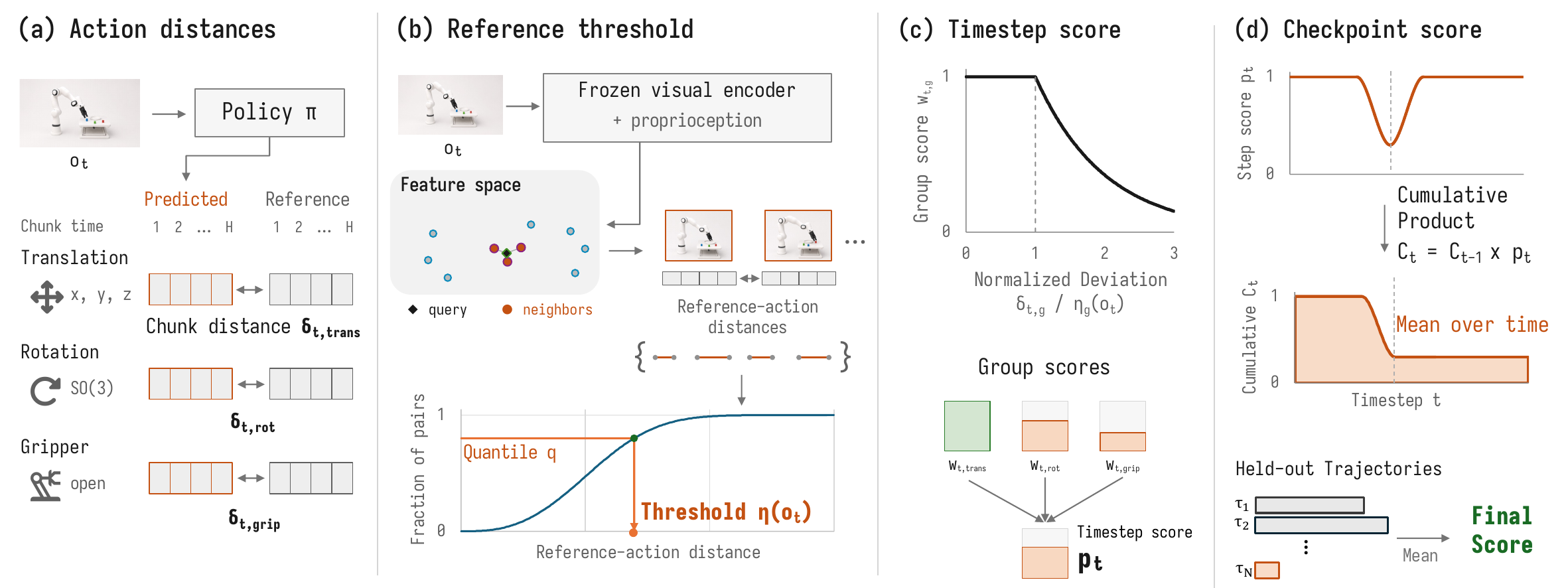}
    \caption{\textbf{\method scoring procedure.} (a) We compare predicted and reference action chunks within each action group to obtain $\delta_{t,g}$. (b) Distances between eligible pairs of reference chunks define a distribution whose quantile at level $q$ gives the state-conditional threshold $\eta_g(o_t)$. The representation, neighborhoods, and thresholds are shared across candidate checkpoints. Both comparisons use the same group distance $D_g$; the illustrated configuration averages the largest $50\%$ of within-chunk timestep distances in physical units. (c) Threshold-normalized deviations yield group scores, which a soft minimum combines into $p_t\in[0,1]$. (d) We form running products of the timestep scores, then average over time and across trajectories to rank checkpoints; higher scores are preferred.}
    \label{figure:method}
\end{figure}

    \subsection{\method}
        \label{sec:method:surval}
        \method evaluates checkpoints in two stages (\Cref{figure:method}). First, state-conditional calibration compares action deviations against thresholds derived from reference actions at similar states, yielding timestep scores in $[0,1]$ (\Cref{subsubsec:scoring_action_deviations}). Second, temporal aggregation forms running products of these scores, so that a low score reduces the contribution of later reference predictions. Averaging the resulting curves over time and across trajectories gives the checkpoint score (\Cref{subsubsec:checkpoint_scores}).

      \subsubsection{State-Conditional Calibration}
      \label{subsubsec:scoring_action_deviations}

      We first describe how to assign a timestep score \(p_t\in[0,1]\) to a single predicted action chunk. At observation \(o_t\), let \(\mathbf a_t=(a_t,\ldots,a_{t+H-1})\) denote the reference action chunk of length \(H\), and let \(\hat{\mathbf a}_t\) denote a chunk predicted by the policy.

      \textbf{Group distances}\quad As action components differ in physical units and geometry, computing a single Euclidean distance over the full action vector can therefore overemphasize components with larger numerical scales and misrepresent rotational differences. To resolve this, we first restore actions to their physical units and partition them into \(G\) groups, such as translation and rotation. We compute distances separately for each group, using Euclidean distance for Euclidean components and the geodesic angle on \(\mathrm{SO}(3)\) for rotations. For chunked predictions, we aggregate these distances over the prediction horizon. Averaging all timestep distances can dilute a brief but large deviation. We therefore define \(D_g\) as the mean of the largest \(50\%\) of the timestep distances within group \(g\), giving greater weight to larger deviations. The predicted chunk's deviation in group \(g\) is defined as
      \begin{equation*}
        \delta_{t,g} = D_g\!\left(\hat{\mathbf a}_t,\mathbf a_t\right).
      \end{equation*}

      \textbf{Reference thresholds}\quad To determine which deviations to penalize, we use differences between actions taken at similar states in successful reference trajectories. Our design treats deviations within a chosen range of these differences as acceptable and penalizes larger deviations. Because this range can vary across states, we introduce state-conditioned threshold: for each observation \(o_t\), we retrieve the \(k\) nearest reference observations in \(\mathcal D_{\mathrm{val}}\) using frozen visual features and proprioception, then compute pairwise distances between their action chunks using the same \(D_g\).\footnote{For joint-velocity control, pairs are restricted to the same reference trajectory; for task-space control, pairs may also span trajectories.}  We set \(\eta_g(o_t)\) to the \(q\)-quantile of these distances. The quantile level \(q\) controls the coverage of this reference range: larger values yield higher thresholds and penalize only larger deviations. The resulting threshold is tighter where nearby reference actions are more consistent (\textit{e.g.} fine-grained placement) and looser where they differ more (\textit{e.g.,} free-space movement).

      \textbf{Timestep scores}\quad To compare deviations across action groups on a common relative scale, we divide each group deviation by its own threshold. Deviations within the threshold receive a score of one, so closer imitation within this range does not improve the score. Beyond the threshold, an exponential decay penalizes the relative excess:
      \begin{equation*}
      w_{t,g}
      =
      \exp\!\left(
      -\left[
      \frac{\delta_{t,g}}{\eta_g(o_t)}-1
      \right]_+
      \right),
      \end{equation*}
      where \([x]_+=\max(x,0)\).

      We combine the group scores using a normalized soft minimum. This aggregation is symmetric across groups while giving greater influence to lower scores from any group:
      \begin{equation*}
      p_t
      =
      -\log\!\left(
      \frac{1}{G}\sum_{g=1}^{G}\exp\!\left(-w_{t,g}\right)
      \right).
      \end{equation*}
      We can also evaluate \(M\) action chunks predicted at the same observation. We score each chunk using the same reference and thresholds, then average the resulting scores as \(p_t = \frac{1}{M}\sum_{m=1}^{M}p_t^{(m)}\), where \(p_t^{(m)}\) is the score of the \(m\)-th chunk. Scoring each chunk before averaging preserves the effect of deviations above the threshold that averaging raw distances may conceal. Empirically, averaging over multiple chunks yields more stable scores and improves checkpoint-ranking accuracy compared with a single sample.

      \subsubsection{Temporal Aggregation}
      \label{subsubsec:checkpoint_scores}
      To reduce the contribution of later predictions after substantial deviations, we form the running product of the timestep scores. For trajectory \(\tau^{(j)}\), we initialize \(C_0^{(j)}=1\) and define
      \begin{equation*}
      C_t^{(j)} = C_{t-1}^{(j)}p_t^{(j)} = \prod_{u=1}^{t}p_u^{(j)}.
      \end{equation*}
      A timestep score of one leaves the cumulative value unchanged, whereas a lower score reduces it from that point onward. Repeated low scores compound this reduction.
      We score each trajectory by its mean cumulative score over the \(T_j\) evaluated timesteps, equivalently the normalized area under the cumulative score curve (AUC):
      \begin{equation*}
      S_{\tau}^{(j)}
      = \frac{1}{T_j}\sum_{t=1}^{T_j} C_t^{(j)}.
      \end{equation*}
      An earlier decline lowers more terms in this mean, while the preceding trajectory prefix retains its contribution.

      Finally, we get the final score $S(\pi;\mathcal D_{\mathrm{val}})$ by averaging the trajectory scores over the held-out set:
      \begin{equation*}
      S(\pi;\mathcal D_{\mathrm{val}})
      = \frac{1}{N_{\mathrm{val}}}
      \sum_{j=1}^{N_{\mathrm{val}}} S_{\tau}^{(j)}.
      \end{equation*}
      For the candidate set \(\Pi\), we compute \(\{S(\pi;\mathcal D_{\mathrm{val}})\}_{\pi\in\Pi}\) and rank checkpoints in descending order of their scores.

\par
\providecommand{\res}[3]{#1{\tiny\,[#2,#3]}}
\providecommand{\resbest}[3]{\textbf{#1}{\tiny\,[#2,#3]}}
\begin{table}[!t]
\centering
\caption{\textbf{Rollout-free checkpoint ranking on DexMimicGen.} All criteria utilizes the same $10$ successful trajectories per source to rank $12$ checkpoints per run. Entries are means with $95\%$ hierarchical bootstrap intervals, averaged over $10$ training seeds per task and then eight tasks. Arrows indicate better values; bold marks the best mean per metric within each source. See \Cref{sec:exp} for the evaluation protocol.}\label{tab:ranking_results}
\footnotesize
\setlength{\tabcolsep}{3pt}
\begin{tabular}{llcccccc}
\toprule
Validation source & Criterion & Hit@1$\uparrow$ & Hit@3$\uparrow$ & Hit@5$\uparrow$ & $\rho\uparrow$ & MMRV$\downarrow$ & nRegret$\downarrow$ \\
\midrule
\multirow{4}{*}{Held-out Demos}
 & \cellcolor{green!12}\method & \cellcolor{green!12}\resbest{0.44}{0.23}{0.65} & \cellcolor{green!12}\cellcolor{green!12}\resbest{0.75}{0.59}{0.90} & \cellcolor{green!12}\resbest{0.95}{0.88}{1.00} & \cellcolor{green!12}\resbest{0.83}{0.72}{0.92} & \cellcolor{green!12}\resbest{0.030}{0.012}{0.053} & \cellcolor{green!12}\resbest{0.082}{0.035}{0.142} \\
 & Validation Loss & \res{0.31}{0.12}{0.53} & \res{0.59}{0.33}{0.82} & \res{0.75}{0.49}{0.97} & \res{0.52}{0.01}{0.93} & \res{0.042}{0.011}{0.088} & \res{0.238}{0.030}{0.504} \\
 & Validation MSE & \res{0.00}{0.00}{0.00} & \res{0.08}{0.00}{0.17} & \res{0.20}{0.04}{0.40} & \res{\textminus{}0.22}{\textminus{}0.62}{0.20} & \res{0.122}{0.068}{0.180} & \res{0.495}{0.325}{0.660} \\
 & Off-Manifold Norm & \res{0.25}{0.11}{0.40} & \res{0.54}{0.33}{0.74} & \res{0.72}{0.46}{0.94} & \res{0.60}{0.22}{0.87} & \res{0.042}{0.022}{0.067} & \res{0.146}{0.052}{0.288} \\
\midrule
\multirow{4}{*}{Rollouts}
 & \cellcolor{green!12}\method & \cellcolor{green!12}\resbest{0.27}{0.12}{0.45} & \cellcolor{green!12}\resbest{0.62}{0.38}{0.83} & \cellcolor{green!12}\resbest{0.90}{0.77}{1.00} & \cellcolor{green!12}\resbest{0.79}{0.69}{0.88} & \cellcolor{green!12}\resbest{0.034}{0.021}{0.047} & \cellcolor{green!12}\resbest{0.113}{0.049}{0.186} \\
 & Validation Loss & \res{0.03}{0.00}{0.10} & \res{0.12}{0.02}{0.27} & \res{0.15}{0.03}{0.30} & \res{\textminus{}0.10}{\textminus{}0.45}{0.25} & \res{0.136}{0.086}{0.190} & \res{0.402}{0.298}{0.514} \\
 & Validation MSE & \res{0.08}{0.02}{0.18} & \res{0.22}{0.05}{0.40} & \res{0.30}{0.12}{0.50} & \res{\textminus{}0.06}{\textminus{}0.27}{0.16} & \res{0.129}{0.084}{0.177} & \res{0.480}{0.343}{0.633} \\
 & Off-Manifold Norm & \res{0.17}{0.07}{0.27} & \res{0.37}{0.23}{0.50} & \res{0.55}{0.42}{0.68} & \res{0.53}{0.39}{0.66} & \res{0.068}{0.047}{0.094} & \res{0.164}{0.123}{0.207} \\
\midrule
\multirow{4}{*}{Noised Rollouts}
 & \cellcolor{green!12}\method & \cellcolor{green!12}\resbest{0.32}{0.19}{0.49} & \cellcolor{green!12}\resbest{0.69}{0.52}{0.85} & \cellcolor{green!12}\resbest{0.91}{0.81}{0.99} & \cellcolor{green!12}\resbest{0.78}{0.68}{0.86} & \cellcolor{green!12}\resbest{0.035}{0.021}{0.053} & \cellcolor{green!12}\resbest{0.081}{0.053}{0.116} \\
 & Validation Loss & \res{0.00}{0.00}{0.00} & \res{0.04}{0.00}{0.11} & \res{0.09}{0.01}{0.19} & \res{\textminus{}0.47}{\textminus{}0.68}{\textminus{}0.15} & \res{0.154}{0.105}{0.202} & \res{0.613}{0.462}{0.773} \\
 & Validation MSE & \res{0.09}{0.00}{0.21} & \res{0.29}{0.11}{0.49} & \res{0.39}{0.17}{0.61} & \res{0.30}{\textminus{}0.05}{0.61} & \res{0.083}{0.041}{0.130} & \res{0.288}{0.148}{0.443} \\
 & Off-Manifold Norm & \res{0.06}{0.01}{0.14} & \res{0.31}{0.16}{0.47} & \res{0.42}{0.25}{0.59} & \res{0.30}{0.05}{0.52} & \res{0.085}{0.055}{0.122} & \res{0.276}{0.185}{0.405} \\
\bottomrule
\end{tabular}
\end{table}

\section{Experiments}
\label{sec:exp}
We design our experiments to answer three questions:
\begin{itemize}
  \item Can \method rank checkpoints more reliably from held-out demonstrations, and retain useful rankings with other successful reference sources? (\Cref{sec:sim_exp})
  \item Does it support real-robot checkpoint selection across policy families and task demands? (\Cref{sec:real_exp})
  \item Which scoring choices contribute to ranking quality, and how do changes in successful references affect it? (\Cref{sec:ablation})
\end{itemize}

Formal metric definitions and task-level results are provided in \Cref{app:metrics,app:sim_per_task,app:real_per_task}.


\paragraph*{Baselines}
We compare three criteria on held-out demonstrations: \mname{Validation Loss}, the behavior-cloning training objective, \mname{Validation MSE}, the mean squared action prediction error independent of the training objectives, and \mname{Off-Manifold Norm}~\citep{pan2025muchado}, the residual norm after projection onto a local basis formed from the $k$ nearest neighbors. For visibility, we negate all baseline values so higher scores are preferred like our proposed score.

\paragraph*{Metrics}
Let $y_i$ denote the empirical success rate of checkpoint $i$ and $s_i$ its offline criterion score (higher is better). \mname{Spearman correlation $\rho$} measures agreement between the rankings induced by $s_i$ and $y_i$. \mname{Hit@$K$ ($K\in\{1,3,5\}$)} is the fraction of runs in which the empirically best checkpoint is among the criterion's top-$K$ candidates. \mname{nRegret} measures the success-rate gap between the empirically best checkpoint and the criterion's highest-scoring checkpoint, normalized by the best-minus-worst success-rate range.
\mname{Mean Maximum Rank Violation (MMRV)} averages each checkpoint's largest success-rate gap among its pairwise ranking disagreements, assigning zero when none occur. Thus, ranking errors involving larger performance gaps incur greater penalties. We refer to~\citep{li2024evaluating} for the formal definition.


\subsection{Simulation Experiments}
\label{sec:sim_exp}
\paragraph*{Setup}
We evaluate eight DexMimicGen~\citep{jiang2025dexmimicgen} tasks, excluding Transport. The action partition follows the robot embodiment and its control interface. For bimanual Franka with parallel grippers (\emph{Threading} and \emph{Three Piece Assembly}), each arm contributes separate end-effector translation and rotation groups, giving $G=4$, and binary gripper commands are excluded. For bimanual Franka with dexterous hands (\emph{Box Cleanup}, \emph{Lift Tray}, and \emph{Drawer Cleanup}), each hand's absolute joint-position vector forms one additional group, giving $G=6$. For GR-1 (\emph{Pouring}, \emph{Coffee}, and \emph{Can Sort}), we group the absolute joint positions of each arm and hand separately, giving $G=4$. Translation and joint-position vectors use Euclidean distance, while rotations use the geodesic angle on $\mathrm{SO}(3)$.

For each task, we train flow-matching $\mathtt{Chi\text{-}UNet}$ policies~\citep{chi2024diffusion,pan2025muchado} using behavior cloning on $500$ demonstrations following RoboMimic~\citep{mandlekar2021what}, aggregated with $10$ independent training seeds. Each run yields $12$ checkpoints saved every $50$ epochs over $600$ epochs. Each checkpoint's empirical success rate is the mean of its own $200$ deployment rollouts.
To test each metric under varying degrees of distribution shift, we construct a separate validation set of $10$ successful trajectories for each of the following three sources: (i)~\mname{Held-out Demonstrations} are from expert trajectories matching the training distribution exactly but not used for training. (ii)~\mname{Rollouts} denote successful closed-loop trajectories from the separately trained policy, introducing mild shift through the policy's own state visitation, and (iii)~\mname{Noised Rollouts} are collected using successful rollouts from the same policy perturbed with Gaussian action noise of standard deviation $0.02$, clipped to $\pm0.05$, representing the largest shift.

For each reference source, we compute metrics per training run, average the $10$ seeds within each task, and then average the task means with equal weights. We draw $10{,}000$ hierarchical bootstrap resamples: each resample draws tasks with replacement, then training seeds with replacement within each sampled task, and recomputes the same aggregate. The notation $[L,U]$ denotes the lower and upper bounds of the resulting $95\%$ bootstrap interval.

\paragraph*{Results}
As shown in \Cref{tab:ranking_results}, \method achieves the strongest ranking quality across all metrics and configurations. Crucially, \method maintains stable scores as the validation distribution shifts from clean demonstrations to noisy rollouts, whereas baselines like Validation Loss and Off Manifold Norm collapse significantly under distribution shift and exhibit negative correlation values. These results confirm that \method provides a reliable ranking signal regardless of how the validation trajectories are collected, a critical property for real-robot workflows where clean demonstrations are scarce.

\begin{figure}[!t]
    \centering
    \captionsetup[subfigure]{font=small,justification=centering}
    \begin{minipage}[c]{0.2\textwidth}
        \centering
        \begin{subfigure}[b]{\textwidth}
            \includegraphics[width=\textwidth]{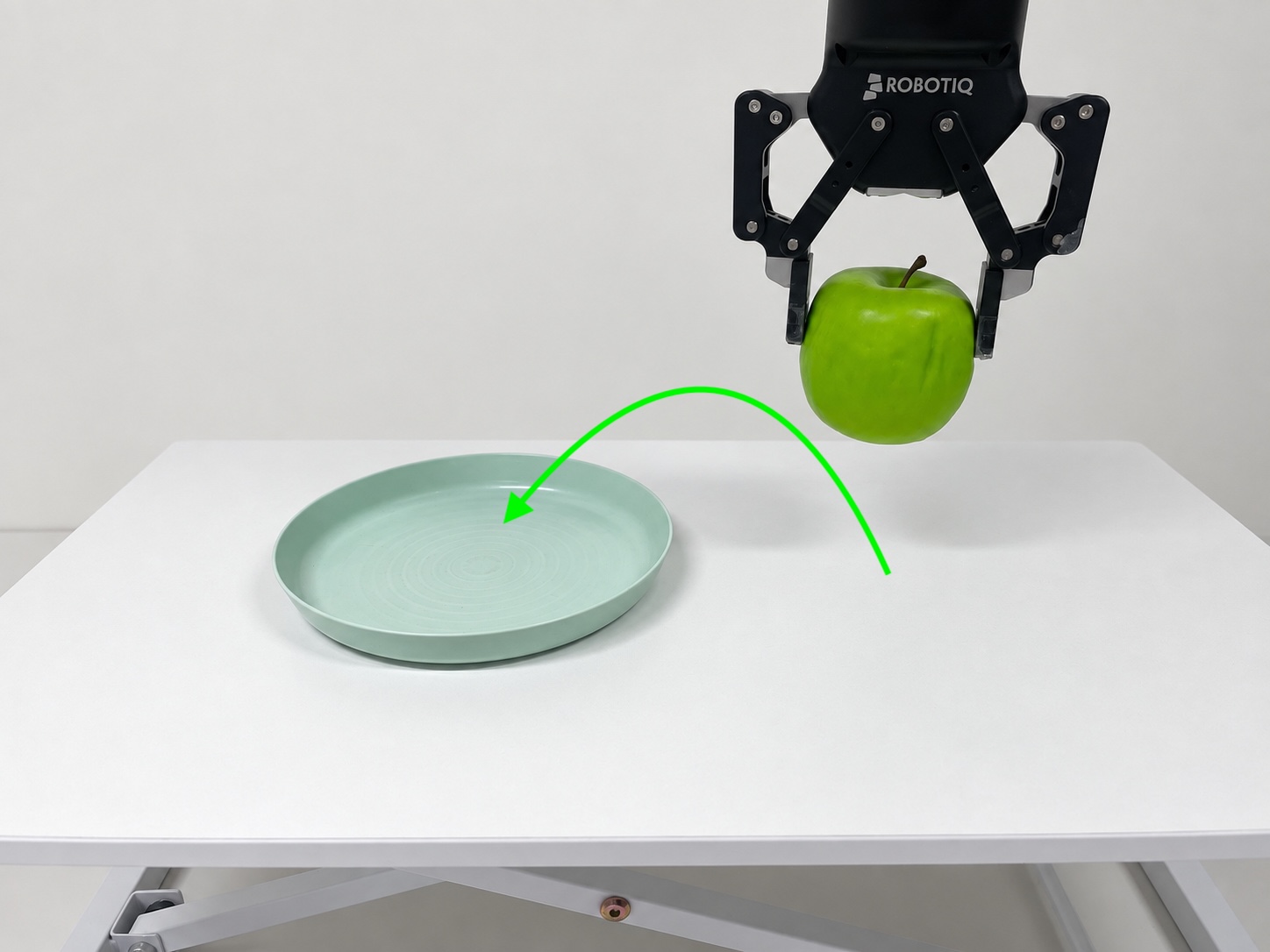}
            \caption{\textit{PnP Apple}}
            \label{fig:task_pnp_apple}
        \end{subfigure}\\[2pt]
        \begin{subfigure}[b]{\textwidth}
            \includegraphics[width=\textwidth]{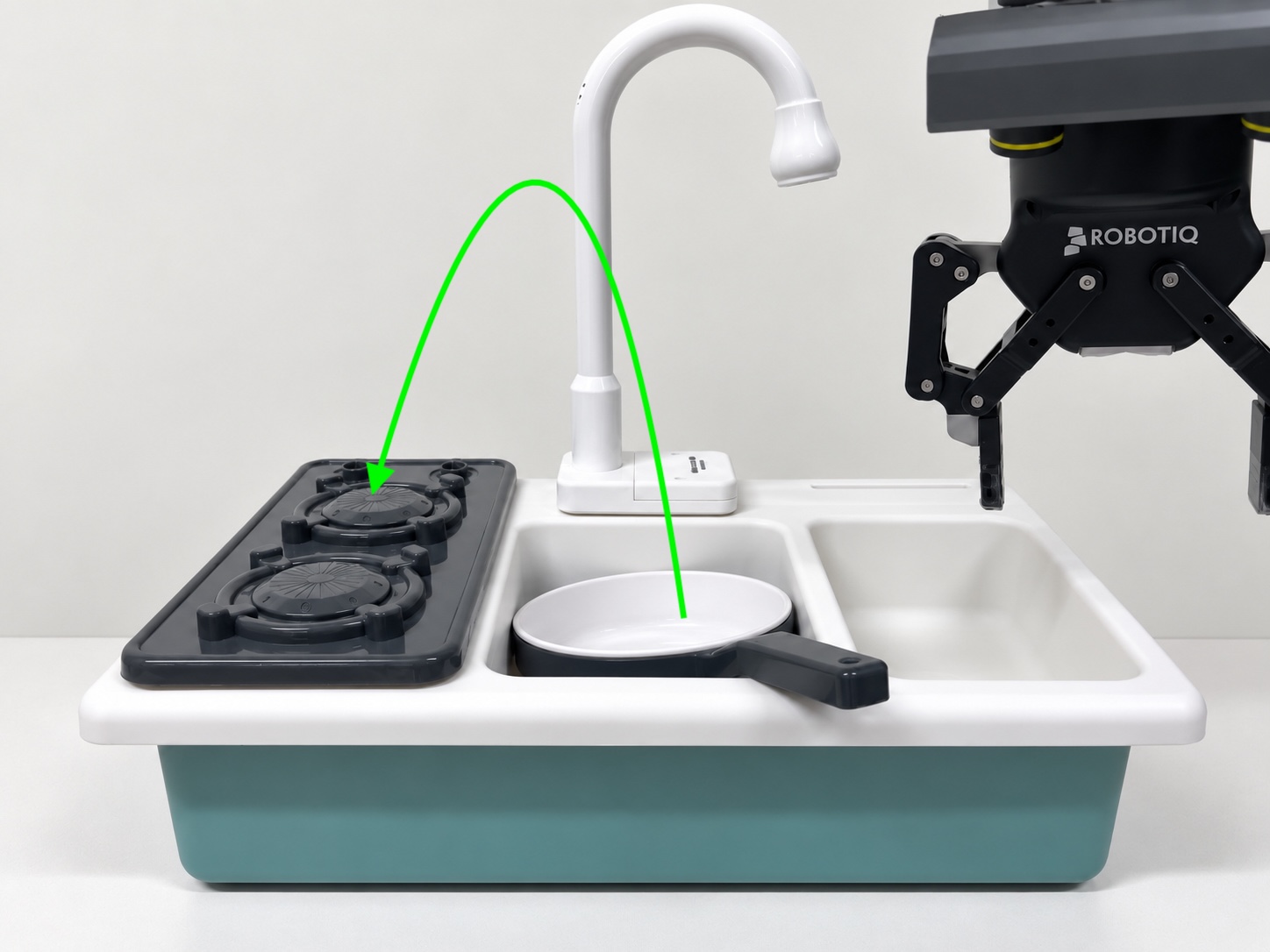}
            \caption{\textit{Put Pan on Stove}}
            \label{fig:task_put_pan_sink}
        \end{subfigure}\\[2pt]
        \begin{subfigure}[b]{\textwidth}
            \includegraphics[width=\textwidth]{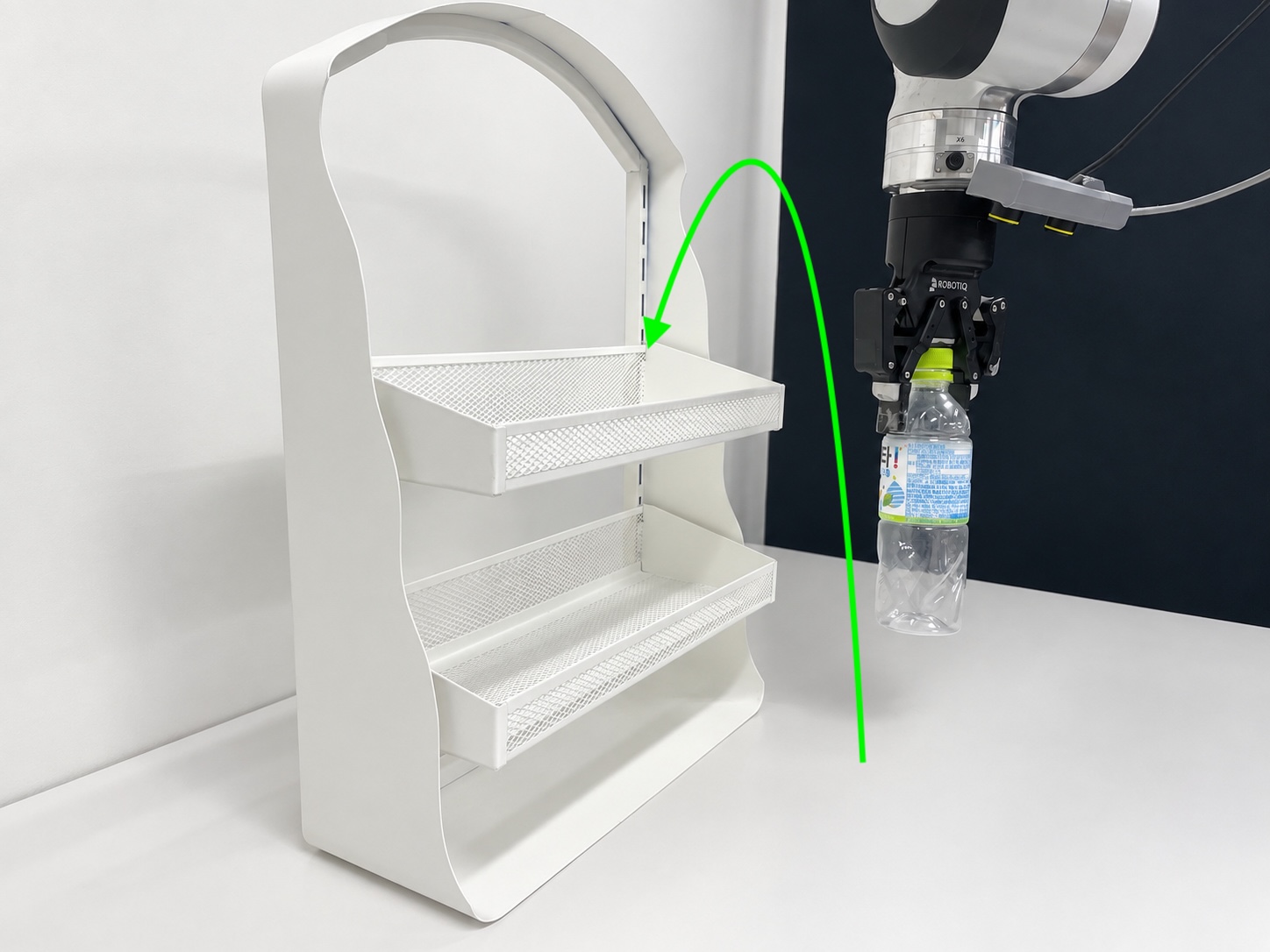}
            \caption{\textit{Put Pet on Shelf}}
            \label{fig:task_put_pet_shelf}
        \end{subfigure}
    \end{minipage}%
    \hspace{0.02\textwidth}%
    \begin{minipage}[c]{0.7\textwidth}
        \centering
        \begin{subfigure}[b]{.49\textwidth}
            \centering
            \includegraphics[width=\textwidth]{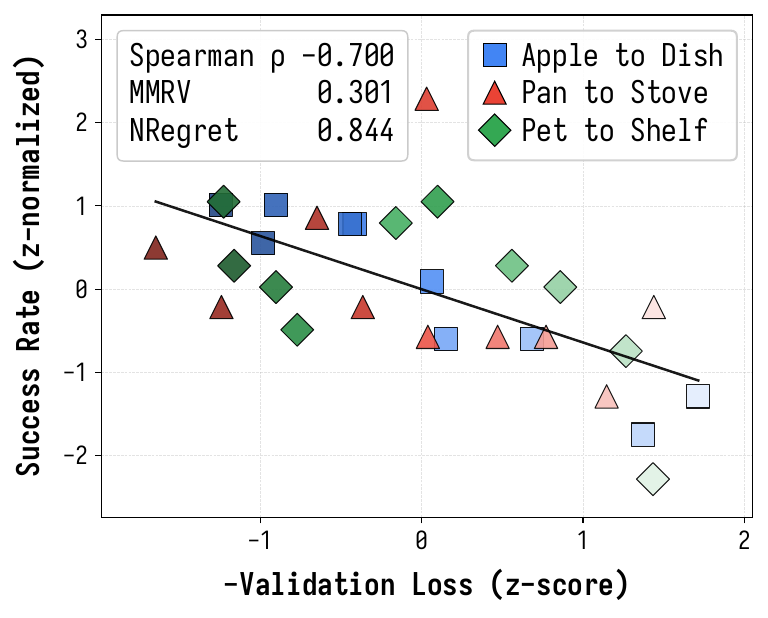}\hfill
        \end{subfigure}
        \begin{subfigure}[b]{.49\textwidth}
            \centering
            \includegraphics[width=\textwidth]{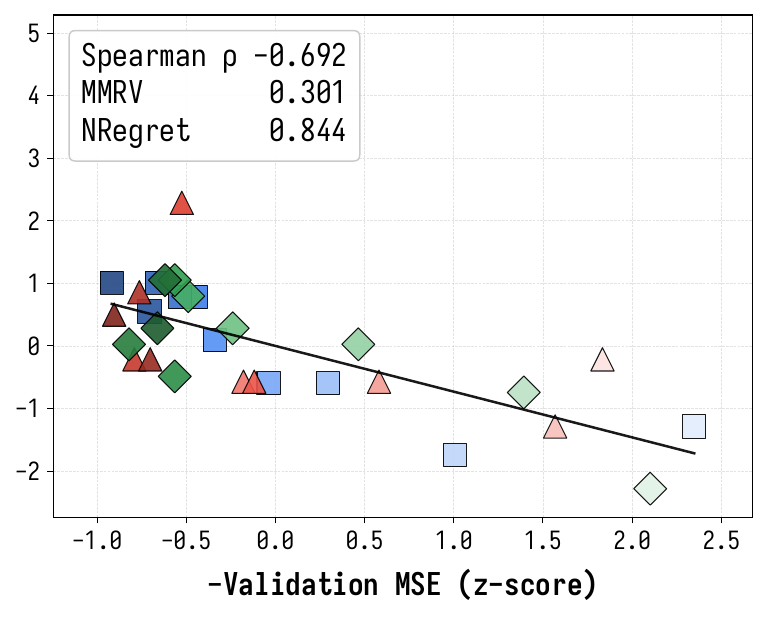}
        \end{subfigure}\\[8pt]
        \begin{subfigure}[b]{.49\textwidth}
            \centering
            \includegraphics[width=\textwidth]{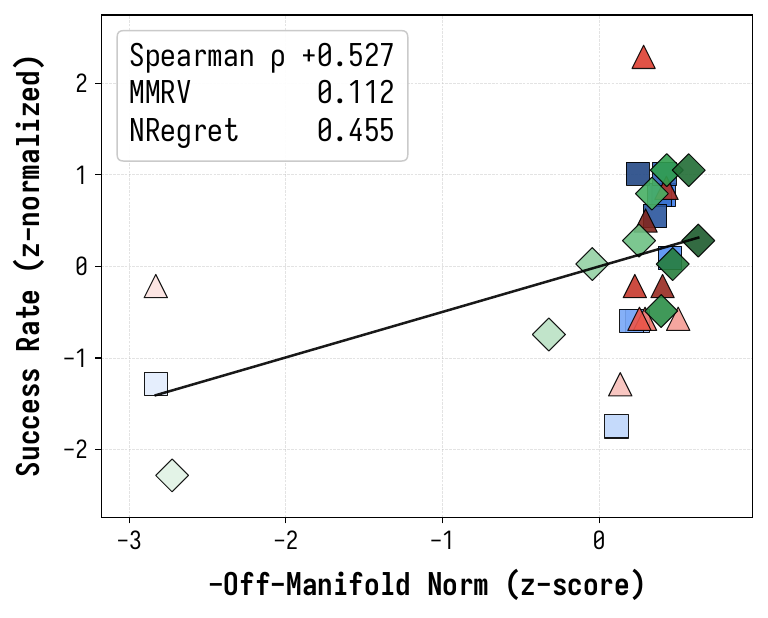}\hfill
        \end{subfigure}
        \begin{subfigure}[b]{.49\textwidth}
            \centering
            \includegraphics[width=\textwidth]{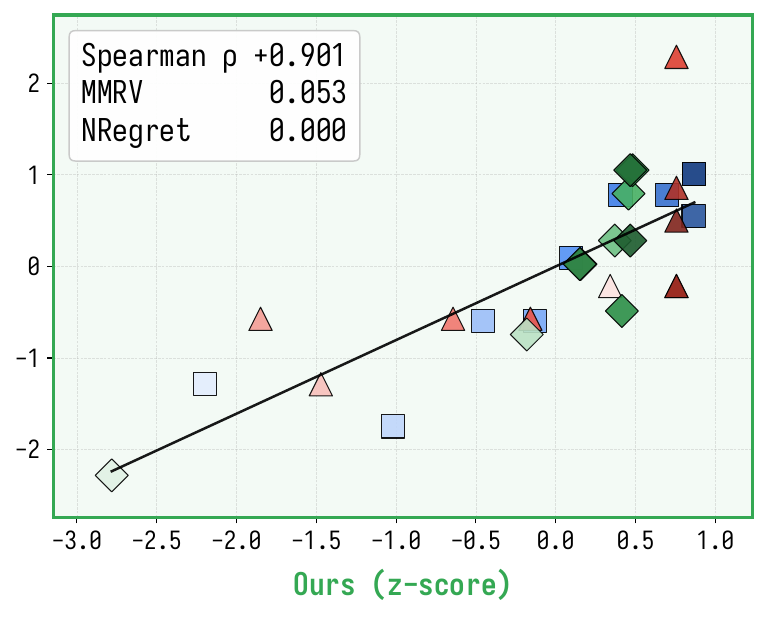}
        \end{subfigure}\\[8pt]
    \end{minipage}
    \caption{\textbf{Diffusion Policy experiments.} (Left) task setups. (Right) z-normalized criterion scores versus z-normalized empirical success rates, with one marker per checkpoint; marker shape and color distinguish tasks. Each task uses $20$ training and $30$ held-out validation demonstrations, $10$ candidate checkpoints, and $30$ deployment rollouts per checkpoint ($900$ rollouts across the three tasks). \method uses task-specific scoring settings. Panel annotations report point estimates of Spearman $\rho\uparrow$, MMRV$\downarrow$, and NRegret$\downarrow$.}
    \label{fig:dp_exp}
\end{figure}

\begin{figure}[!t]
    \centering
    \captionsetup[subfigure]{font=small,justification=centering}
    \begin{minipage}[c]{0.2\textwidth}
        \centering
        \begin{subfigure}[b]{\textwidth}
            \includegraphics[width=\textwidth]{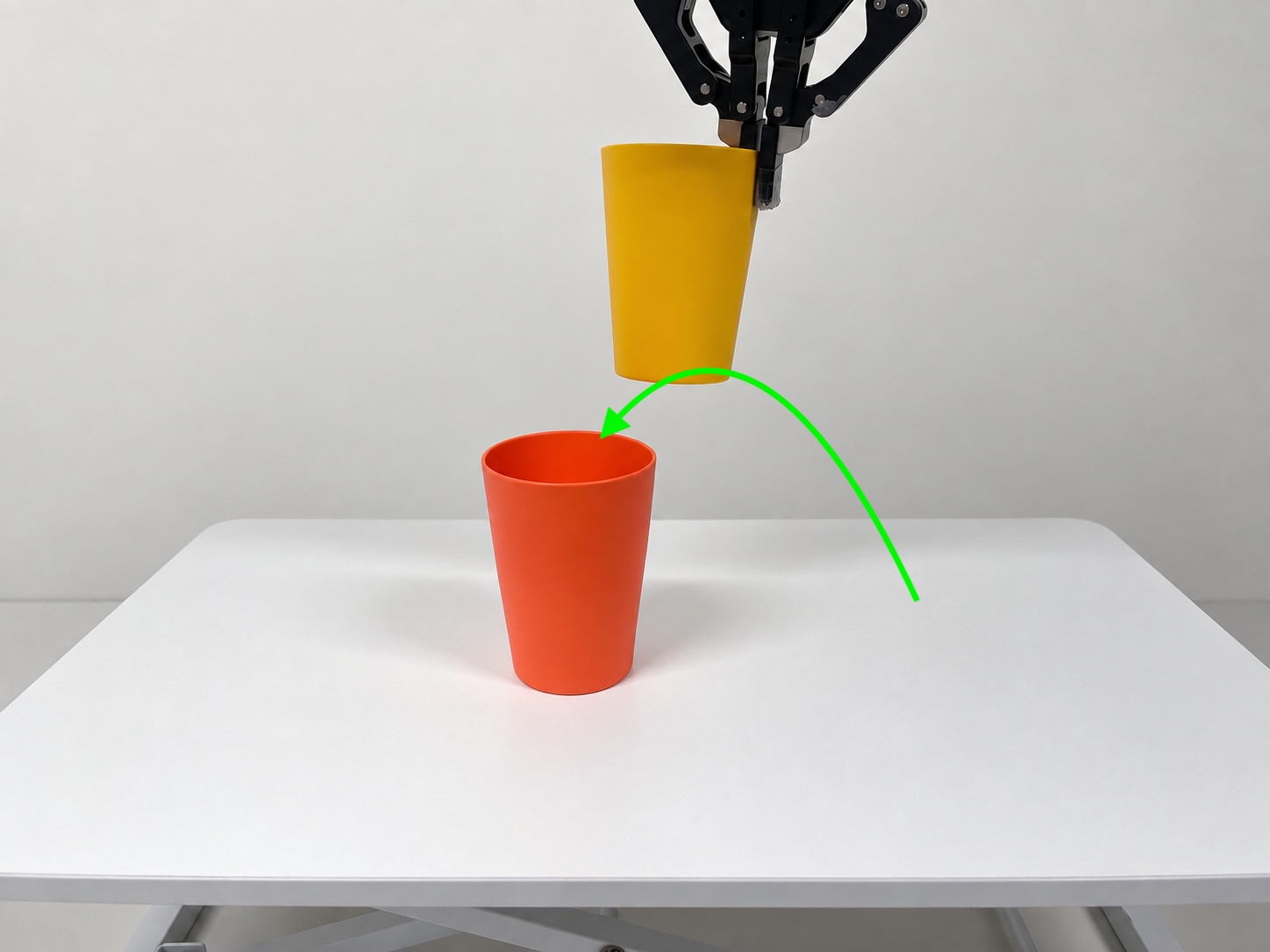}
            \caption{\textit{Nest Cups}}
            \label{fig:task_nest_cups}
        \end{subfigure}\\[2pt]
        \begin{subfigure}[b]{\textwidth}
            \includegraphics[width=\textwidth]{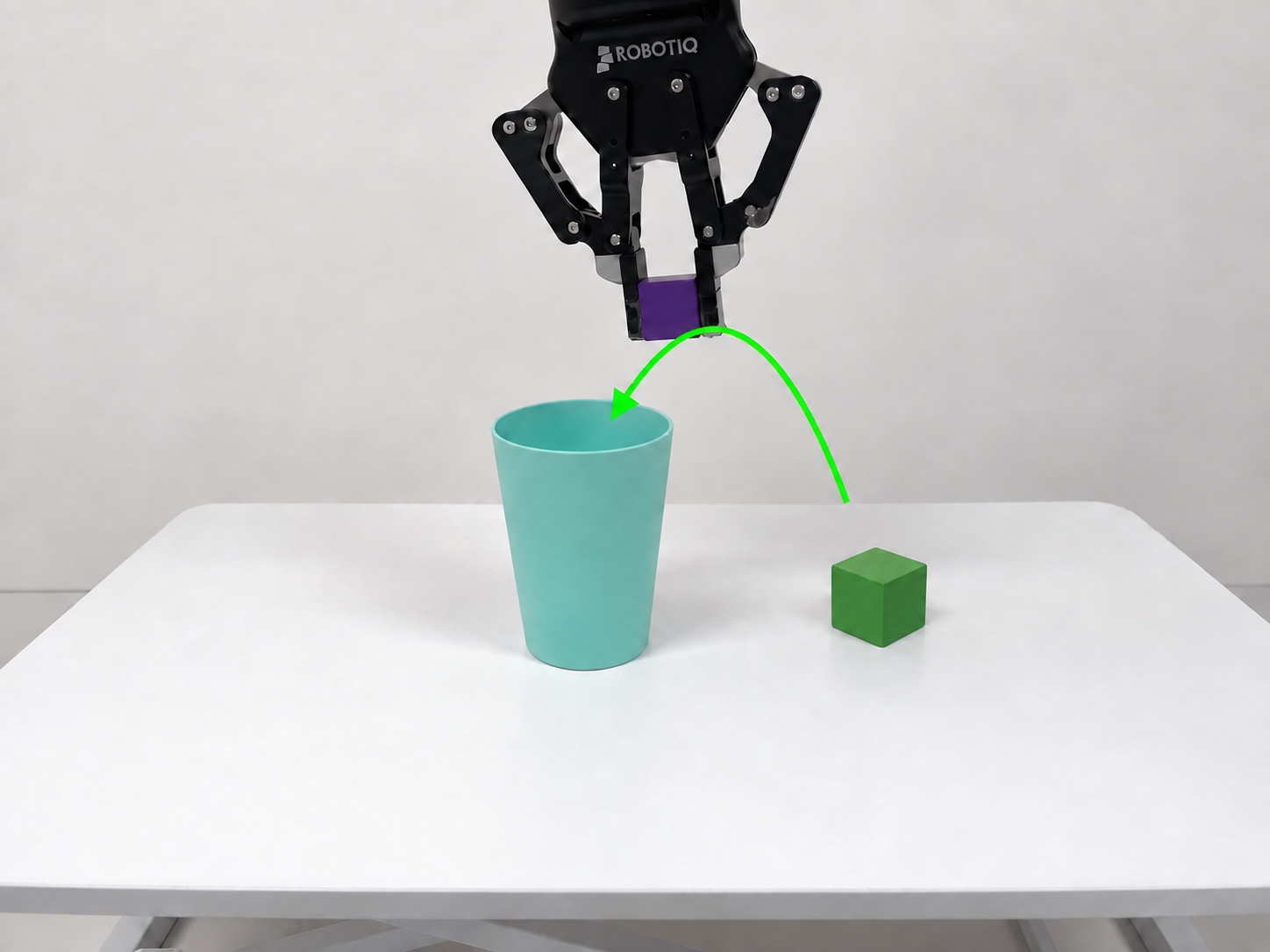}
            \caption{\textit{Put Two Cubes}}
            \label{fig:task_put_two_cubes}
        \end{subfigure}\\[2pt]
        \begin{subfigure}[b]{\textwidth}
            \includegraphics[width=\textwidth]{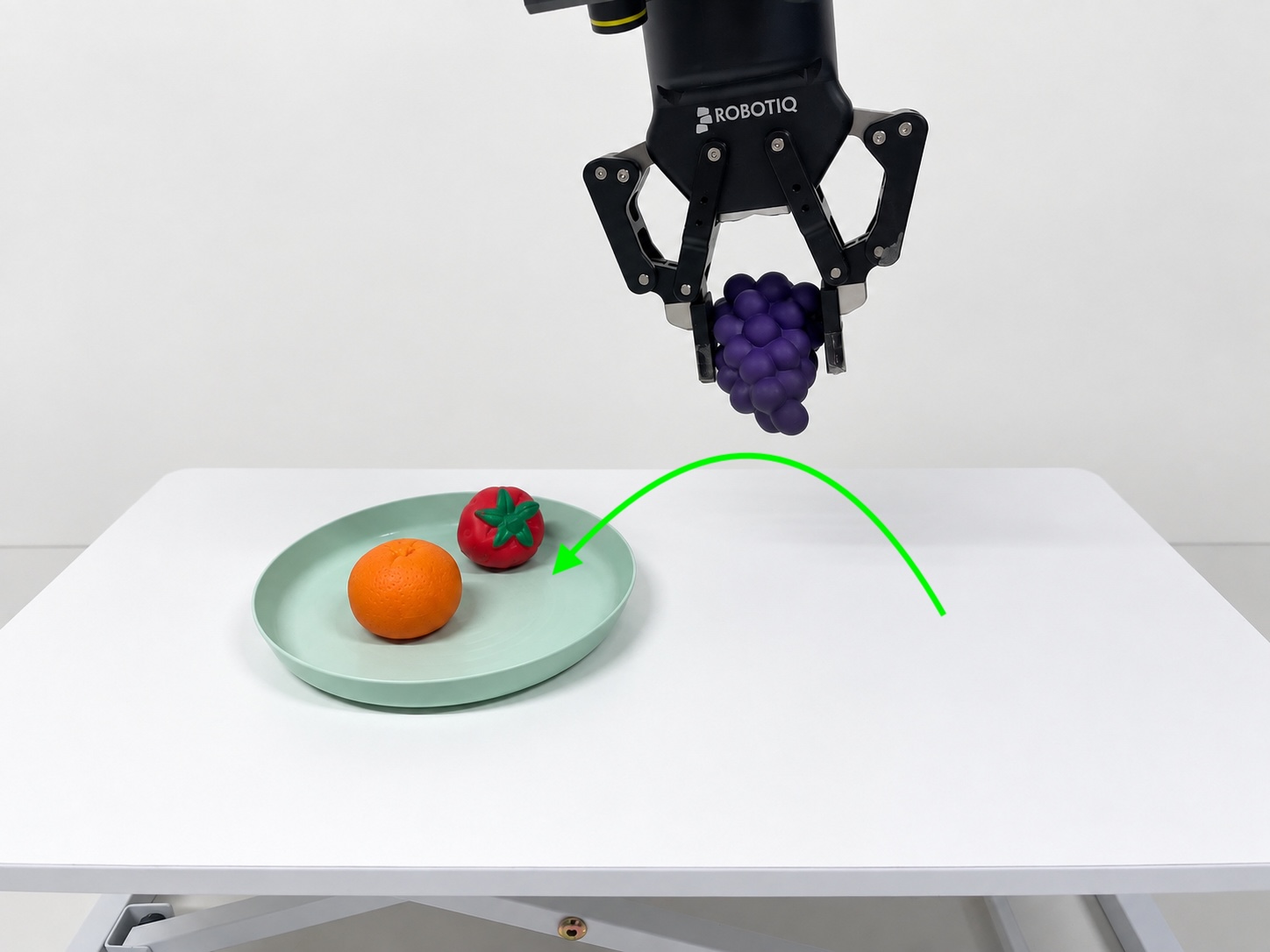}
            \caption{\textit{PnP 3 Fruit}}
            \label{fig:task_pnp_3_fruit}
        \end{subfigure}
    \end{minipage}%
    \hspace{0.02\textwidth}%
    \begin{minipage}[c]{0.7\textwidth}
        \centering
        \begin{subfigure}[b]{.49\textwidth}
            \centering
            \includegraphics[width=\textwidth]{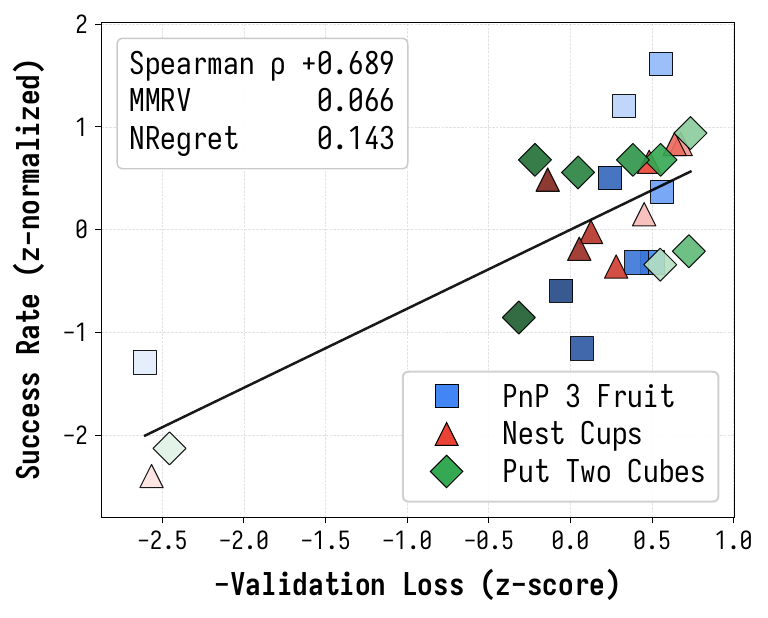}\hfill
        \end{subfigure}
        \begin{subfigure}[b]{.49\textwidth}
            \centering
            \includegraphics[width=\textwidth]{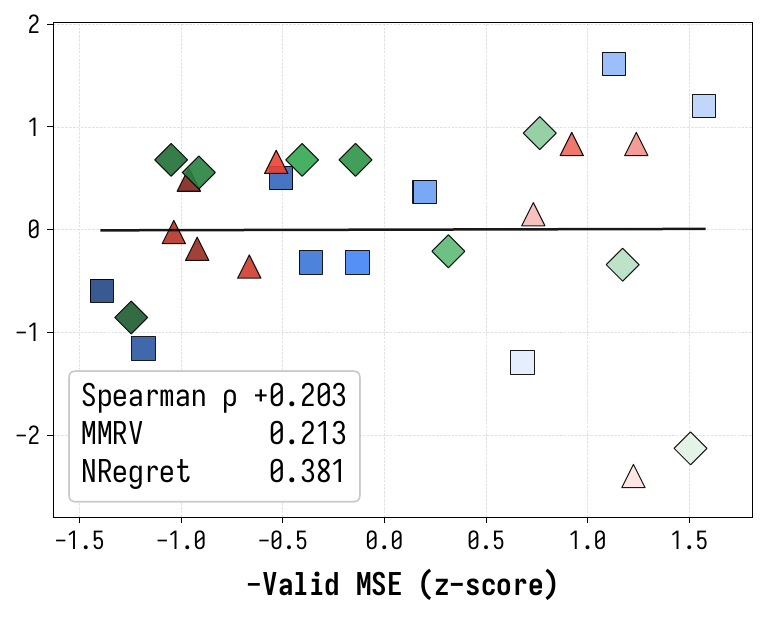}
        \end{subfigure}\\[8pt]
        \begin{subfigure}[b]{.49\textwidth}
            \centering
            \includegraphics[width=\textwidth]{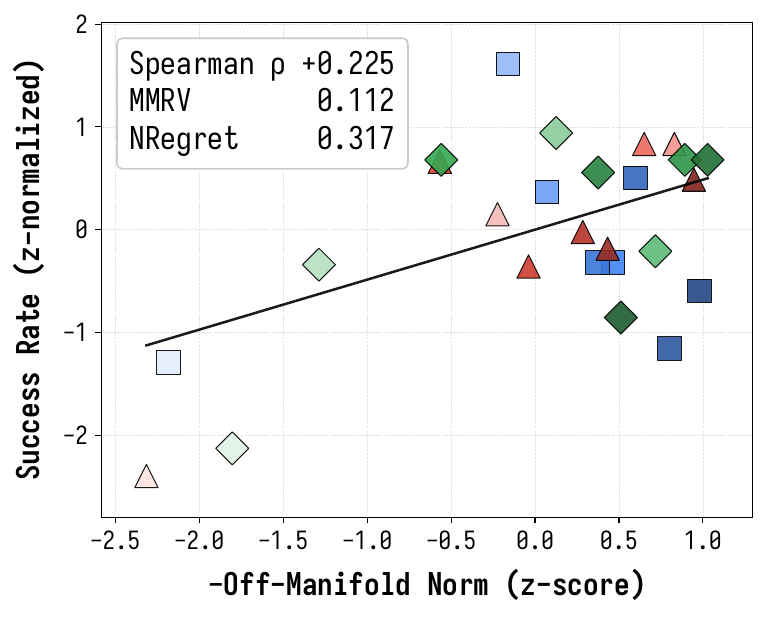}\hfill
        \end{subfigure}
        \begin{subfigure}[b]{.49\textwidth}
            \centering
            \includegraphics[width=\textwidth]{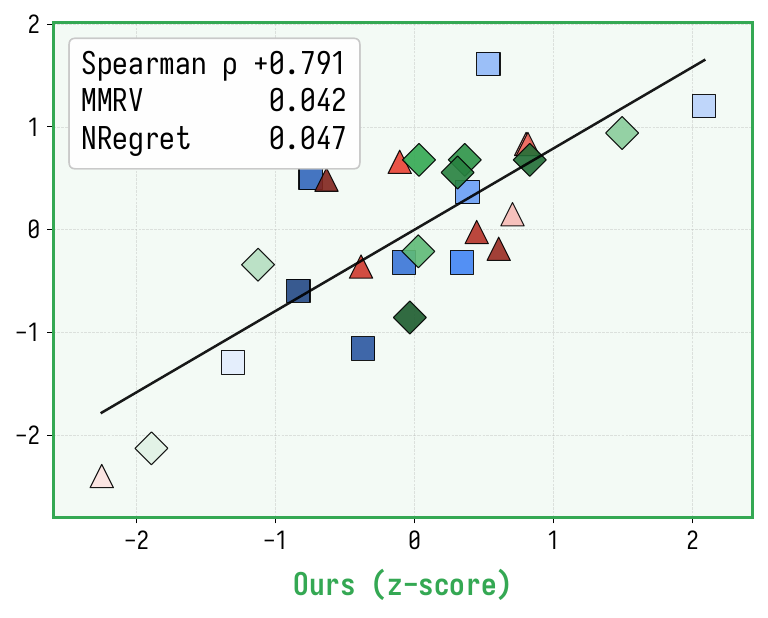}
        \end{subfigure}\\[8pt]
    \end{minipage}
    \caption{\textbf{$\pi_{0.5}$-DROID experiments.} (Left) task setups. (Right) z-normalized criterion scores versus z-normalized empirical success rates, with one marker per checkpoint; marker shape and color distinguish tasks. Policies use joint-velocity control and are fine-tuned on $20$ target-task training demonstrations per task. Each task uses $30$ held-out validation demonstrations, $10$ candidate checkpoints, and $30$ deployment rollouts per checkpoint ($900$ rollouts across the three tasks). \method uses task-specific scoring settings. Panel annotations report point estimates of Spearman $\rho\uparrow$, MMRV$\downarrow$, and NRegret$\downarrow$.}
    \label{fig:vla_exp}
\end{figure}

\subsection{Real-World Experiments}
\label{sec:real_exp}
To validate that \method generalizes beyond simulation, we evaluate it on real-robot manipulation tasks using two policy architectures of different scales: a task-specific diffusion policy trained from scratch and a large pretrained VLA fine-tuned on in-domain data. All experiments use a Franka Research 3 arm following DROID~\citep{khazatsky2024droid} hardware and software setup, with task-specific demonstrations collected via teleoperation. In both settings, held-out demonstrations unseen during training serve as the validation set, and baselines and ranking metrics follow the same protocol as in \Cref{sec:sim_exp}. To support reliable empirical comparison, we evaluate each of $10$ checkpoints with $30$ deployment rollouts on each of three tasks, yielding $30\times10\times3=900$ real-robot evaluations per policy family and $1{,}800$ in total.

\subsubsection*{Diffusion Policy Experiments}
\paragraph*{Setup}
We fine-tune DROID Diffusion Policy~\citep{khazatsky2024droid} with pre-trained ResNet-50 encoder using two exterior left camera views. The policy is co-trained with the DROID dataset for $50$ epochs on $20$ task-specific demonstrations and checkpointed every $5$ epochs. We evaluate on three tabletop tasks---\emph{PnP Apple}, \emph{Put Pan on Stove}, and \emph{Put Pet on Shelf}---each validated against $30$ held-out demonstrations. The policy predicts relative end-effector translation and rotation, with rotations represented in 6D~\citep{zhou2019continuity}, together with a binary gripper command. For scoring, translation and rotation form two separate groups ($G=2$), and the gripper is excluded. We use Euclidean distance for translation and convert the 6D rotation representation to $\mathrm{SO}(3)$ before computing the geodesic angle.

\paragraph*{Results}
As shown in \Cref{fig:dp_exp}, standard baselines struggle in this setting: Validation Loss and Validation MSE produces highly negative correlation values and reverses the true checkpoint ranking. While Off-Manifold Norm recovers moderate correlation, this result is mainly due to the separation of early checkpoints, and all other checkpoints are indistinguishable. In contrast, \method consistently ranks checkpoints in alignment with deployment success across all three tasks, demonstrating its robustness and reliability for real-robot checkpoint selection.

\subsubsection*{VLA Experiments}
\paragraph*{Setup}
To test \method on longer-horizon tasks that require precise sequential manipulation, we fine-tune $\pi_{0.5}$-DROID~\citep{black2025pi} using wrist camera and a single exterior left camera view. The policy uses 7-DoF joint-velocity control. We train and evaluate on 3 tasks with varying difficulties: \emph{Nest Cups}, \emph{Put Two Cubes}, and \emph{PnP 3 Fruit}. All tasks are trained with $20$ demonstrations and validated against $30$ held-out demonstrations. For scoring, each arm joint forms a separate scalar group ($G=7$), with the absolute difference between predicted and reference joint velocities as its distance. The binary gripper command is excluded.

\paragraph*{Results}
As shown in \Cref{fig:vla_exp}, the baselines recover moderate correlation in this setting, yet \method leads by a clear margin. The gap is most pronounced on \emph{Put Two Cubes} and \emph{PnP 3 Fruit}, where longer horizons amplify compounding errors that one-step losses fail to capture. We hypothesize that baselines perform better here than in the Diffusion Policy setting because $\pi_{0.5}$'s large-scale pretraining provides well-structured representations, ensuring that validation loss tracks genuine policy improvement rather than overfitting. Nevertheless, \method still outperforms all baselines by a consistent margin and, across both architectures, reliably selects near-optimal checkpoints on real hardware with irreducible variables from human operators and sensor noise.

\subsection{Analyses and Ablation Studies}
\label{sec:ablation}

\paragraph{Perturbations with similar prediction error}
The motivating example in \Cref{sec:method:motivating_example} raises a practical question: can similar prediction errors accompany different deployment consequences? On \emph{Box Cleanup}, we perturb one trained policy uniformly ($\sigma=0.05$) or at bottlenecks (\textit{i.e.,} robot aligning the box lid to the box) ($\sigma=0.2$), matching average prediction error and evaluating $50$ paired rollouts from identical initial states. Despite similar MSE and Off-Manifold Norm changes, bottleneck perturbations degrade success and \method scores more strongly (\Cref{tab:perturbation}). This supports the complete score's sensitivity to consequential deviations. Success-difference ranges use paired resampling; offline metric changes are point estimates.

\begin{table}[!t]
    \centering
    \caption{\textbf{Perturbation Analysis} We apply different scale of perturbations on trained policies for \emph{Box Cleanup} tasks, and compared changes relative to the original policy using $50$ fixed scenes.}
    \label{tab:perturbation}
    \setlength{\tabcolsep}{6pt}
    \small
    \begin{tabular*}{\linewidth}{@{\extracolsep{\fill}}lrrrr@{}}
        \toprule
        Variant & $\Delta$Success & $\Delta$MSE & $\Delta$Off-Man. & $\Delta$\method \\
        \midrule
        Uniform ($\sigma{=}0.05$) & -4.0\%P & -2.46\% & -2.51\% & -1.42\% \\
        Bottleneck ($\sigma{=}0.2$) & -40.0\%P & -2.47\% & -2.00\% & -52.69\% \\
        \bottomrule
    \end{tabular*}
\end{table}

\providecommand{\abres}[3]{\shortstack[c]{#1\\{\scriptsize[#2,#3]}}}
\providecommand{\abresbest}[3]{\shortstack[c]{\textbf{#1}\\{\scriptsize[#2,#3]}}}

\begin{table}[!t]
\centering
\small
\setlength{\tabcolsep}{3pt}
\renewcommand{\thesubtable}{(\alph{subtable})}
\captionsetup[subtable]{labelformat=simple,font=small,justification=centering,singlelinecheck=true,skip=4pt,hypcap=true}
\caption{\textbf{Component and temporal ablations across four tasks.} (a) Removing calibration threshold uses $q=1.0$ for all groups, and removing state conditioning fixes the threshold across states. (b) Both reductions use identical timestep scores. Means are shown above 95\% hierarchical bootstrap intervals. Bold marks the best value per column within each panel.}
\label{tab:analyses}
\begin{subtable}[t]{0.48\linewidth}
\centering
\caption{Component ablation}
\label{tab:analyses:components}
\begin{tabular*}{\linewidth}{@{\extracolsep{\fill}}lccc@{}}
\toprule
Setting & $\rho\uparrow$ & MMRV$\downarrow$ & nRegret$\downarrow$ \\
\midrule
Full & \abresbest{0.76}{0.66}{0.86} & \abresbest{0.040}{0.02}{0.07} & \abresbest{0.096}{0.05}{0.15} \\
w/o action group & \abres{0.69}{0.56}{0.81} & \abres{0.048}{0.03}{0.07} & \abres{0.148}{0.09}{0.21} \\
w/o calib.\ threshold & \abres{0.31}{0.11}{0.52} & \abres{0.093}{0.05}{0.13} & \abres{0.363}{0.29}{0.44} \\
w/o state cond. & \abres{0.42}{0.20}{0.68} & \abres{0.083}{0.03}{0.14} & \abres{0.334}{0.19}{0.47} \\
\bottomrule
\end{tabular*}
\end{subtable}%
\hfill
\begin{subtable}[t]{0.48\linewidth}
\centering
\caption{Temporal aggregation}
\label{tab:analyses:temporal}
\begin{tabular*}{\linewidth}{@{\extracolsep{\fill}}lccc@{}}
\toprule
Setting & $\rho\uparrow$ & MMRV$\downarrow$ & nRegret$\downarrow$ \\
\midrule
Mean & \abres{0.63}{0.40}{0.78} & \abres{0.052}{0.02}{0.10} & \abres{0.180}{0.09}{0.33} \\
Cumulative scoring & \abresbest{0.76}{0.66}{0.86} & \abresbest{0.040}{0.02}{0.07} & \abresbest{0.096}{0.05}{0.15} \\
\bottomrule
\end{tabular*}
\end{subtable}
\end{table}

\paragraph{Component ablation}
To identify which design choices contribute to ranking, we remove action grouping (Grp), threshold scoring (Thr), or state-conditional adaptation (SC) on \emph{Coffee}, \emph{Drawer Cleanup}, \emph{Pouring}, and \emph{Three Piece Assembly}. Each removal lowers aggregate ranking quality (\Cref{tab:analyses:components}), supporting comparisons that account for heterogeneous action units and calibrate deviations against state-relative reference variation.

\paragraph{Temporal aggregation}
The perturbation comparison evaluates the complete score; to separate temporal aggregation from calibration, we compare the per-trajectory mean of identical stored $p_t$ values with cumulative scoring (\Cref{subsubsec:checkpoint_scores}), which averages their running products over time. Both methods then average trajectory scores with equal weights. Candidates, references, thresholds, sampled predictions, and deployment outcomes are fixed. Across four DexMimicGen tasks, cumulative scoring improves ranking correlation and lowers selection regret compared with the mean (\Cref{tab:analyses:temporal}). This supports reducing the contribution of later reference predictions after a substantial deviation, beyond calibrating individual timestep scores.

\begin{table}[!t]
\centering
\small
\setlength{\tabcolsep}{6pt}
\renewcommand{\abres}[3]{#1\,{\scriptsize[#2,#3]}}
\renewcommand{\abresbest}[3]{\textbf{#1}\,{\scriptsize[#2,#3]}}
\caption{\textbf{Task-specific versus shared settings.} Means over $24$ task/reference-source combinations across eight tasks, with 95\% hierarchical bootstrap intervals in brackets. Shared (task $k$) varies only neighborhood size $k$. Bold marks the best value per column.}
\label{tab:analyses:shared}
\begin{tabular*}{\linewidth}{@{\extracolsep{\fill}}lccc@{}}
\toprule
Setting & $\rho\uparrow$ & MMRV$\downarrow$ & nRegret$\downarrow$ \\
\midrule
Tuned & \abresbest{0.80}{0.74}{0.86} & \abresbest{0.033}{0.02}{0.05} & \abresbest{0.090}{0.06}{0.12} \\
Shared (task $k$) & \abres{0.75}{0.67}{0.81} & \abres{0.037}{0.03}{0.05} & \abres{0.110}{0.08}{0.15} \\
Shared & \abres{0.69}{0.57}{0.78} & \abres{0.038}{0.03}{0.05} & \abres{0.135}{0.10}{0.17} \\
\bottomrule
\end{tabular*}
\end{table}

\paragraph{Shared scoring settings}
To assess tuning burden, we compare task-specific (Tuned), fully Shared, and Shared settings with task-specific neighborhood size $k$ across the $22$ task/reference-source combinations spanning the eight DexMimicGen tasks. Fully Shared settings reduce ranking quality relative to Tuned (\Cref{tab:analyses:shared}). Allowing only $k$ to vary by task recovers part of the gap, indicating that adapting neighborhood size provides some of the benefit of task-specific tuning while sharing the remaining settings.

\paragraph{Augmenting with rollouts}
On \textit{PnP Apple}, supplementing held-out demonstrations with successful final-policy rollouts improves checkpoint ranking (\Cref{tab:analyses:real_rollouts}). This suggests that successful policy rollouts can complement held-out demonstrations as reference data, improving selection without changing the scoring rule.

\begin{table}[!t]
\centering
\small
\setlength{\tabcolsep}{6pt}
\renewcommand{\thesubtable}{(\alph{subtable})}
\captionsetup[subtable]{labelformat=simple,font=small,justification=centering,singlelinecheck=true,skip=4pt,hypcap=true}
\caption{\textbf{Real-world reference-set analyses.} (a) Adding successful final-policy rollouts to held-out demonstrations on \textit{PnP Apple}. (b) Using successful policy rollouts containing a random $3$--$5$ second pause. $\rho$ denotes Spearman correlation in (a) and Pearson correlation in (b). Values are point estimates.}
\label{tab:analyses:real_references}
\begin{subtable}[t]{0.48\linewidth}
\centering
\caption{Rollout augmentation}
\label{tab:analyses:real_rollouts}
\begin{tabular*}{\linewidth}{@{\extracolsep{\fill}}lcc@{}}
\toprule
Reference set & $\rho\uparrow$ & MMRV$\downarrow$ \\
\midrule
Held-out & $0.834$ & $0.230$ \\
+ Rollouts & \textbf{0.920} & \textbf{0.070} \\
\bottomrule
\end{tabular*}
\end{subtable}%
\hfill
\begin{subtable}[t]{0.48\linewidth}
\centering
\caption{Temporal artifacts}
\label{tab:analyses:real_paused}
\begin{tabular*}{\linewidth}{@{\extracolsep{\fill}}lcc@{}}
\toprule
Reference set & $\rho\uparrow$ & MMRV$\downarrow$ \\
\midrule
Held-out & $0.834$ & $0.230$ \\
Paused Rollouts & $0.801$ & $0.194$ \\
\bottomrule
\end{tabular*}
\end{subtable}
\end{table}

\paragraph{Robustness to temporal artifacts}
To examine reference trajectories containing idle segments, we collect successful policy rollouts with a random $3$--$5$ second pause at a randomly chosen point in each rollout (\Cref{tab:analyses:real_paused}). Checkpoint rankings remain comparable to those obtained with held-out demonstrations, with slightly lower correlation but lower MMRV. These results suggest that successful references containing idle segments can still provide useful information for checkpoint selection.

\section{Conclusion}
    \label{sec:conclusion}
    We present \method, a novel rollout-free checkpoint selection metric that converts action deviations from successful reference trajectories into a normalized score. \method provides a practical way to compare robot policy checkpoints before deployment using a reusable set of successful trajectories. Our simulation and real-robot results show that \method rankings align more closely with task success than those based on averaged prediction error. This can support more informed decisions during policy development, from tracking progress across checkpoints to focusing limited robot evaluation on promising candidates.

    \looseness=-1
    Despite of its advantages, \method has two limitations. First, even though action variation at nearby successful states is a proxy for tolerable deviation, but it cannot establish whether a candidate policy would recover from an error or predict its behavior at unseen states induced by its own actions. Second, \method provides a relative ranking, not a calibrated success-rate estimate, and selected candidates still require deployment validation.

\bibliographystyle{plainnat}
\bibliography{references}
\clearpage
\appendix
\begin{center}
    {\bfseries\LARGE Appendix}
\end{center}

\section{Experiment Details}
\label{app:sim_setup}

\subsection{Ranking Metrics}
\label{app:metrics}
Let $y_i$ be the empirical success rate and $s_i$ the offline score of checkpoint $i$, with higher scores preferred. For each run, let $i^\star=\arg\max_i y_i$ and $\hat{i}=\arg\max_i s_i$. We break selection ties by checkpoint order, preferring the earliest checkpoint. The set $\operatorname{TopK}(s)$ contains the $K$ highest-scoring checkpoints under the same rule.

\paragraph{Hit@$K$}
The per-run selection indicator is
\begin{equation}
    \operatorname{Hit@K}
    = \mathbf{1}\!\left[i^\star\in\operatorname{TopK}(s)\right].
\end{equation}
Averaging this indicator across runs gives the frequency with which the criterion's shortlist contains the empirically best checkpoint.

\paragraph{Spearman correlation}
Let $r(s)$ and $r(y)$ denote the checkpoint ranks induced by the offline scores and empirical success rates, assigning average ranks to ties. Spearman correlation is
\begin{equation}
    \rho=\operatorname{Corr}\!\left(r(s),r(y)\right),
\end{equation}
where $\operatorname{Corr}$ is Pearson correlation. This measures agreement over the full checkpoint ordering rather than only the selected checkpoint.

\paragraph{Mean Maximum Rank Violation}
We compute MMRV~\citep{li2024evaluating} using success-rate gaps:
\begin{equation}
    \operatorname{MMRV}
    = \frac{1}{N}\sum_{i=1}^{N}
    \max_{1\leq j\leq N}
    \left\{
        |y_i-y_j|\,
        \mathbf{1}\!\left[(s_i<s_j)\ne(y_i<y_j)\right]
    \right\}.
\end{equation}
Here $N$ is the number of candidate checkpoints in the run. A checkpoint with no pairwise disagreement contributes zero; a disagreement across a larger success-rate gap contributes a larger penalty.

\paragraph{Normalized regret}
For a nonzero success-rate range, the normalized selection regret is
\begin{equation}
    \operatorname{nRegret}
    = \frac{y_{i^\star}-y_{\hat{i}}}
    {\max_i y_i-\min_i y_i}.
\end{equation}
This dimensionless quantity measures the selected checkpoint's performance loss relative to the observed best-to-worst range.

\subsection{Implementation Details}
\label{app:implementation}

\paragraph{Shared reference calibration}
The visual representation used for neighborhood retrieval is frozen, and the reference features, neighborhoods, and thresholds are computed once and shared across candidate checkpoints. Thus, a checkpoint's score changes through its predicted actions rather than through a changing reference representation or calibration. \Cref{alg:race,alg:threshold} specify the computation and its reuse. The group distance $D_g$ follows \Cref{subsubsec:scoring_action_deviations}, and all action distances are computed in physical units. The pseudocode assumes that each retrieved neighborhood supplies at least one eligible pair of action chunks and that the resulting thresholds satisfy $\eta_g(o_t)>0$.

\begin{algorithm}[H]
\caption{\method checkpoint scoring}
\label{alg:race}
\begin{algorithmic}[1]
\REQUIRE Candidate checkpoints $\Pi$; successful references $\mathcal D_{\mathrm{val}}$ with actions in physical units; frozen retrieval representation $\psi$; group distances $D_1,\ldots,D_G$; chunk length $H$; neighborhood size $k$; quantile level $q$; number of samples $M$
\ENSURE A score $S(\pi;\mathcal D_{\mathrm{val}})$ for each $\pi\in\Pi$
\STATE Compute shared thresholds $\eta_g(o_t^{(j)})$ using \Cref{alg:threshold}
\FOR{each checkpoint $\pi\in\Pi$}
    \FOR{each reference trajectory $\tau^{(j)}\in\mathcal D_{\mathrm{val}}$}
        \STATE $C_0^{(j)}\gets1$
        \FOR{each evaluated observation $o_t^{(j)}$, $t=1,\ldots,T_j$}
            \STATE Read the reference action chunk $\mathbf a_t^{(j)}$ of length $H$
            \FOR{$m=1,\ldots,M$}
                \STATE Sample $\hat{\mathbf a}_t^{(j,m)}\sim\pi(o_t^{(j)})$ and restore physical units
                \FOR{$g=1,\ldots,G$}
                    \STATE $\delta_{t,g}^{(j,m)}\gets D_g(\hat{\mathbf a}_t^{(j,m)},\mathbf a_t^{(j)})$
                    \STATE $w_{t,g}^{(j,m)}\gets\exp\!\left(-[\delta_{t,g}^{(j,m)}/\eta_g(o_t^{(j)})-1]_+\right)$
                \ENDFOR
                \STATE $p_t^{(j,m)}\gets-\log\!\left(G^{-1}\sum_{g=1}^{G}\exp(-w_{t,g}^{(j,m)})\right)$
            \ENDFOR
            \STATE $p_t^{(j)}\gets M^{-1}\sum_{m=1}^{M}p_t^{(j,m)}$
            \STATE $C_t^{(j)}\gets C_{t-1}^{(j)}p_t^{(j)}$
        \ENDFOR
        \STATE $S_\tau^{(j)}\gets T_j^{-1}\sum_{t=1}^{T_j}C_t^{(j)}$
    \ENDFOR
    \STATE $S(\pi;\mathcal D_{\mathrm{val}})\gets N_{\mathrm{val}}^{-1}\sum_{j=1}^{N_{\mathrm{val}}}S_\tau^{(j)}$
\ENDFOR
\end{algorithmic}
\end{algorithm}

\begin{algorithm}[H]
\caption{State-conditional reference thresholds}
\label{alg:threshold}
\begin{algorithmic}[1]
\REQUIRE Successful references $\mathcal D_{\mathrm{val}}$ with action chunks of length $H$ in physical units; frozen retrieval representation $\psi$; neighborhood size $k$; quantile level $q$; group distances $D_1,\ldots,D_G$
\ENSURE Thresholds $\eta_g(o_t^{(j)})$ for all evaluated reference observations and groups
\STATE Encode reference observations using $\psi$, combining frozen visual features and proprioception
\FOR{each evaluated reference observation $o_t^{(j)}$}
    \STATE Retrieve its $k$ nearest reference observations and their action chunks
    \STATE Form eligible pairs $\mathcal P_t^{(j)}$ of distinct retrieved chunks: same-trajectory pairs for joint-velocity control; within- or across-trajectory pairs for task-space control
    \FOR{$g=1,\ldots,G$}
        \STATE $\mathcal E_{t,g}^{(j)}\gets\{D_g(\mathbf a_u,\mathbf a_v):(u,v)\in\mathcal P_t^{(j)}\}$
        \STATE $\eta_g(o_t^{(j)})\gets\operatorname{Quantile}_q(\mathcal E_{t,g}^{(j)})$
    \ENDFOR
\ENDFOR
\end{algorithmic}
\end{algorithm}

\paragraph{Off-Manifold Norm}
For a query state, the baseline retrieves nearby expert states and stacks their actions into a local basis $A=[a^{(1)},\ldots,a^{(k)}]$. For a predicted action $a$, its residual norm is
\begin{equation}
    \operatorname{OMN}(a)=\lVert a-P_A(a)\rVert_2,
\end{equation}
where $P_A$ projects onto the span of the columns of $A$~\citep{pan2025muchado}. The offline ranking uses the negative residual. In the Diffusion Policy experiments, Off-Manifold Norm retrieves references using the same frozen policy encoder as \method. The $\pi_{0.5}$-DROID experiments use the original Off-Manifold Norm baseline.

\paragraph{Real-world data and software}
The Diffusion Policy experiments use the released DROID policy-learning code\footnote{\url{https://github.com/droid-dataset/droid_policy_learning}}. Training mixes DROID data and target-task demonstrations in equal proportions, and observation and action normalization uses the DROID dataset statistics. The $\pi_{0.5}$-DROID experiments use the released pretrained checkpoint and the RLDS dataset fine-tuning recipe from OpenPI\footnote{\url{https://github.com/Physical-Intelligence/openpi}}. The control interfaces, action groups, camera inputs, and task sets are specified in \Cref{sec:real_exp}.

\subsubsection{Task success criteria and initialization}
\noindent\textbf{Diffusion Policy tasks}
\begin{itemize}
    \item \textit{PnP Apple}: The robot grasps an apple from the table and places it in a bowl. Success requires the apple to come to rest fully inside the bowl; its initial position is randomized within a $5\times5$~cm tabletop region.
    \item \textit{Put Pet on Shelf}: The robot places a plastic bottle on the top shelf. Success requires the bottle to rest stably on the shelf surface; its initial position is randomized within a $5\times5$~cm tabletop region.
\end{itemize}

\noindent\textbf{$\pi_{0.5}$-DROID tasks}
\begin{itemize}
    \item \textit{Nest Cups}: The robot nests an orange cup on the left into a yellow cup on the right. Success requires the orange cup to be fully seated inside the yellow cup; both initial cup positions are randomized within the workspace.
    \item \textit{Put Two Cubes}: The robot picks up two cubes from the table and places them in a cup. Success requires both cubes to rest inside the cup; at episode initialization, the two cubes are randomly assigned to two fixed tabletop positions.
    \item \textit{PnP 3 Fruit}: The robot picks up three toy fruits from the table and places them on a plate. Success requires all three fruits to rest on the plate; each fruit has a fixed nominal initial position, with its placement randomized within a local box of area $2$~cm$^2$ at that position.
\end{itemize}

\clearpage
\section{Task-Level Experimental Results}
\label{app:exps}

\subsection{Simulation Experiments}
\label{app:sim_per_task}
\Cref{tab:per_task_box_cleanup,tab:per_task_can_sort,tab:per_task_coffee,tab:per_task_drawer_cleanup,tab:per_task_lift_tray,tab:per_task_pouring,tab:per_task_threading,tab:per_task_three_piece_assembly} report ranking performance for each DexMimicGen task and reference source, averaged over the training seeds. The strongest criterion depends on the task and reference source: for example, Validation Loss has lower reported regret on held-out demonstrations for \emph{Drawer Cleanup}, whereas \method has lower reported regret for both rollout-based reference sources on that task. The final column lists the reported checkpoint-selection regret.


\begin{table}[!htbp]
\centering
\small
\setlength{\tabcolsep}{3pt}
\caption{\textbf{Box Cleanup.} Per-task ranking performance on DexMimicGen~\citep{jiang2025dexmimicgen}, averaged over $10$ seeds. Arrows indicate better values; bold marks the best reported value per metric within each validation source.}
\label{tab:per_task_box_cleanup}
\begin{tabular*}{\linewidth}{@{\extracolsep{\fill}}llcccccc}
\toprule
Validation source & Criterion & Hit@1$\uparrow$ & Hit@3$\uparrow$ & Hit@5$\uparrow$ & $\rho\uparrow$ & MMRV$\downarrow$ & Regret$\downarrow$ \\
\midrule
\multirow{4}{*}{Held-out Demos} & \cellcolor{green!12}\method & \cellcolor{green!12}\textbf{0.600} & \cellcolor{green!12}\textbf{1.000} & \cellcolor{green!12}\textbf{1.000} & \cellcolor{green!12}\textbf{0.969} & \cellcolor{green!12}\textbf{0.008} & \cellcolor{green!12}\textbf{0.009} \\
 & Validation Loss & 0.400 & 0.800 & \textbf{1.000} & 0.938 & 0.011 & 0.017 \\
 & Validation MSE & 0.000 & 0.000 & 0.000 & -0.892 & 0.256 & 0.754 \\
 & Off-Manifold Norm & 0.300 & 0.700 & 0.900 & 0.848 & 0.028 & 0.048 \\
\midrule
\multirow{4}{*}{Rollouts} & \cellcolor{green!12}\method & \cellcolor{green!12}\textbf{0.300} & \cellcolor{green!12}\textbf{0.800} & \cellcolor{green!12}\textbf{1.000} & \cellcolor{green!12}\textbf{0.902} & \cellcolor{green!12}\textbf{0.016} & \cellcolor{green!12}\textbf{0.041} \\
 & Validation Loss & 0.000 & 0.000 & 0.000 & -0.708 & 0.245 & 0.619 \\
 & Validation MSE & 0.100 & 0.300 & 0.400 & -0.143 & 0.228 & 0.781 \\
 & Off-Manifold Norm & 0.100 & 0.300 & 0.400 & 0.287 & 0.121 & 0.190 \\
\midrule
\multirow{4}{*}{Noised Rollouts} & \cellcolor{green!12}\method & \cellcolor{green!12}\textbf{0.100} & \cellcolor{green!12}\textbf{0.700} & \cellcolor{green!12}\textbf{0.900} & \cellcolor{green!12}\textbf{0.903} & \cellcolor{green!12}\textbf{0.017} & \cellcolor{green!12}\textbf{0.052} \\
 & Validation Loss & 0.000 & 0.000 & 0.200 & -0.655 & 0.247 & 0.569 \\
 & Validation MSE & \textbf{0.100} & 0.600 & 0.600 & 0.573 & 0.052 & 0.118 \\
 & Off-Manifold Norm & 0.000 & 0.600 & 0.700 & 0.413 & 0.097 & 0.207 \\
\bottomrule
\end{tabular*}
\end{table}

\begin{table}[!htbp]
\centering
\small
\setlength{\tabcolsep}{3pt}
\caption{\textbf{Can Sort.} Per-task ranking performance on DexMimicGen~\citep{jiang2025dexmimicgen}, averaged over $10$ seeds. Arrows indicate better values; bold marks the best reported value per metric within each validation source.}
\label{tab:per_task_can_sort}
\begin{tabular*}{\linewidth}{@{\extracolsep{\fill}}llcccccc}
\toprule
Validation source & Criterion & Hit@1$\uparrow$ & Hit@3$\uparrow$ & Hit@5$\uparrow$ & $\rho\uparrow$ & MMRV$\downarrow$ & Regret$\downarrow$ \\
\midrule
\multirow{4}{*}{Held-out Demos} & \cellcolor{green!12}\method & \cellcolor{green!12}\textbf{0.300} & \cellcolor{green!12}\textbf{0.600} & \cellcolor{green!12}\textbf{1.000} & \cellcolor{green!12}\textbf{0.815} & \cellcolor{green!12}\textbf{0.017} & \cellcolor{green!12}\textbf{0.037} \\
 & Validation Loss & 0.000 & 0.000 & 0.100 & -0.886 & 0.073 & 0.981 \\
 & Validation MSE & 0.000 & 0.100 & 0.300 & -0.555 & 0.058 & 0.730 \\
 & Off-Manifold Norm & 0.000 & 0.000 & 0.000 & -0.638 & 0.069 & 0.574 \\
\midrule
\multirow{4}{*}{Rollouts} & \cellcolor{green!12}\method & \cellcolor{green!12}\textbf{0.600} & \cellcolor{green!12}\textbf{0.900} & \cellcolor{green!12}\textbf{1.000} & \cellcolor{green!12}\textbf{0.908} & \cellcolor{green!12}\textbf{0.040} & \cellcolor{green!12}\textbf{0.016} \\
 & Validation Loss & 0.000 & 0.000 & 0.100 & -0.517 & 0.198 & 0.396 \\
 & Validation MSE & 0.100 & 0.500 & 0.600 & 0.200 & 0.133 & 0.311 \\
 & Off-Manifold Norm & 0.200 & 0.400 & 0.600 & 0.461 & 0.086 & 0.110 \\
\midrule
\multirow{4}{*}{Noised Rollouts} & \cellcolor{green!12}\method & \cellcolor{green!12}\textbf{0.700} & \cellcolor{green!12}\textbf{0.900} & \cellcolor{green!12}\textbf{1.000} & \cellcolor{green!12}0.708 & \cellcolor{green!12}0.023 & \cellcolor{green!12}0.063 \\
 & Validation Loss & 0.000 & 0.000 & 0.000 & -0.793 & 0.072 & 0.907 \\
 & Validation MSE & 0.400 & 0.700 & 0.700 & \textbf{0.832} & \textbf{0.008} & \textbf{0.038} \\
 & Off-Manifold Norm & 0.100 & 0.100 & 0.400 & 0.184 & 0.039 & 0.220 \\
\bottomrule
\end{tabular*}
\end{table}
\clearpage

\begin{table}[!htbp]
\centering
\small
\setlength{\tabcolsep}{3pt}
\caption{\textbf{Coffee.} Per-task ranking performance on DexMimicGen~\citep{jiang2025dexmimicgen}, averaged over $10$ seeds. Arrows indicate better values; bold marks the best reported value per metric within each validation source.}
\label{tab:per_task_coffee}
\begin{tabular*}{\linewidth}{@{\extracolsep{\fill}}llcccccc}
\toprule
Validation source & Criterion & Hit@1$\uparrow$ & Hit@3$\uparrow$ & Hit@5$\uparrow$ & $\rho\uparrow$ & MMRV$\downarrow$ & Regret$\downarrow$ \\
\midrule
\multirow{4}{*}{Held-out Demos} & \cellcolor{green!12}\method & \cellcolor{green!12}\textbf{0.100} & \cellcolor{green!12}\textbf{0.400} & \cellcolor{green!12}\textbf{0.800} & \cellcolor{green!12}0.589 & \cellcolor{green!12}0.076 & \cellcolor{green!12}0.184 \\
 & Validation Loss & 0.000 & 0.100 & 0.300 & -0.396 & 0.190 & 0.698 \\
 & Validation MSE & 0.000 & 0.300 & 0.700 & \textbf{0.718} & \textbf{0.045} & \textbf{0.139} \\
 & Off-Manifold Norm & \textbf{0.100} & 0.300 & 0.400 & 0.350 & 0.118 & 0.204 \\
\midrule
\multirow{4}{*}{Rollouts} & \cellcolor{green!12}\method & \cellcolor{green!12}0.100 & \cellcolor{green!12}\textbf{0.300} & \cellcolor{green!12}\textbf{0.800} & \cellcolor{green!12}\textbf{0.727} & \cellcolor{green!12}\textbf{0.032} & \cellcolor{green!12}\textbf{0.154} \\
 & Validation Loss & 0.000 & 0.100 & 0.300 & 0.238 & 0.092 & 0.366 \\
 & Validation MSE & 0.000 & 0.000 & 0.000 & -0.364 & 0.121 & 0.532 \\
 & Off-Manifold Norm & \textbf{0.200} & \textbf{0.300} & 0.600 & 0.663 & 0.047 & 0.173 \\
\midrule
\multirow{4}{*}{Noised Rollouts} & \cellcolor{green!12}\method & \cellcolor{green!12}\textbf{0.400} & \cellcolor{green!12}\textbf{0.500} & \cellcolor{green!12}\textbf{0.800} & \cellcolor{green!12}\textbf{0.770} & \cellcolor{green!12}0.045 & \cellcolor{green!12}\textbf{0.083} \\
 & Validation Loss & 0.000 & 0.000 & 0.000 & -0.719 & 0.205 & 0.984 \\
 & Validation MSE & 0.100 & 0.300 & 0.500 & 0.734 & \textbf{0.041} & 0.164 \\
 & Off-Manifold Norm & 0.000 & 0.000 & 0.000 & -0.408 & 0.197 & 0.645 \\
\bottomrule
\end{tabular*}
\end{table}

\begin{table}[!htbp]
\centering
\small
\setlength{\tabcolsep}{3pt}
\caption{\textbf{Drawer Cleanup.} Per-task ranking performance on DexMimicGen~\citep{jiang2025dexmimicgen}, averaged over $10$ seeds. Arrows indicate better values; bold marks the best reported value per metric within each validation source.}
\label{tab:per_task_drawer_cleanup}
\begin{tabular*}{\linewidth}{@{\extracolsep{\fill}}llcccccc}
\toprule
Validation source & Criterion & Hit@1$\uparrow$ & Hit@3$\uparrow$ & Hit@5$\uparrow$ & $\rho\uparrow$ & MMRV$\downarrow$ & Regret$\downarrow$ \\
\midrule
\multirow{4}{*}{Held-out Demos} & \cellcolor{green!12}\method & \cellcolor{green!12}0.500 & \cellcolor{green!12}0.700 & \cellcolor{green!12}\textbf{1.000} & \cellcolor{green!12}0.639 & \cellcolor{green!12}0.083 & \cellcolor{green!12}0.172 \\
 & Validation Loss & \textbf{0.700} & \textbf{0.900} & \textbf{1.000} & \textbf{0.974} & \textbf{0.006} & \textbf{0.010} \\
 & Validation MSE & 0.000 & 0.000 & 0.000 & -0.895 & 0.223 & 0.645 \\
 & Off-Manifold Norm & 0.400 & 0.800 & 0.900 & 0.900 & 0.024 & 0.025 \\
\midrule
\multirow{4}{*}{Rollouts} & \cellcolor{green!12}\method & \cellcolor{green!12}\textbf{0.600} & \cellcolor{green!12}\textbf{0.900} & \cellcolor{green!12}\textbf{1.000} & \cellcolor{green!12}\textbf{0.908} & \cellcolor{green!12}\textbf{0.040} & \cellcolor{green!12}\textbf{0.016} \\
 & Validation Loss & 0.000 & 0.000 & 0.100 & -0.534 & 0.198 & 0.390 \\
 & Validation MSE & 0.100 & 0.500 & 0.600 & 0.200 & 0.133 & 0.348 \\
 & Off-Manifold Norm & 0.200 & 0.400 & 0.600 & 0.461 & 0.086 & 0.110 \\
\midrule
\multirow{4}{*}{Noised Rollouts} & \cellcolor{green!12}\method & \cellcolor{green!12}\textbf{0.200} & \cellcolor{green!12}\textbf{0.600} & \cellcolor{green!12}\textbf{0.800} & \cellcolor{green!12}\textbf{0.566} & \cellcolor{green!12}\textbf{0.083} & \cellcolor{green!12}\textbf{0.111} \\
 & Validation Loss & 0.000 & 0.000 & 0.000 & -0.638 & 0.205 & 0.480 \\
 & Validation MSE & 0.000 & 0.000 & 0.000 & -0.617 & 0.195 & 0.674 \\
 & Off-Manifold Norm & 0.000 & 0.300 & 0.400 & 0.258 & 0.100 & 0.303 \\
\bottomrule
\end{tabular*}
\end{table}
\clearpage

\begin{table}[!htbp]
\centering
\small
\setlength{\tabcolsep}{3pt}
\caption{\textbf{Lift Tray.} Per-task ranking performance on DexMimicGen~\citep{jiang2025dexmimicgen}, averaged over $10$ seeds. Arrows indicate better values; bold marks the best reported value per metric within each validation source.}
\label{tab:per_task_lift_tray}
\begin{tabular*}{\linewidth}{@{\extracolsep{\fill}}llcccccc}
\toprule
Validation source & Criterion & Hit@1$\uparrow$ & Hit@3$\uparrow$ & Hit@5$\uparrow$ & $\rho\uparrow$ & MMRV$\downarrow$ & Regret$\downarrow$ \\
\midrule
\multirow{4}{*}{Held-out Demos} & \cellcolor{green!12}\method & \cellcolor{green!12}0.600 & \cellcolor{green!12}\textbf{1.000} & \cellcolor{green!12}\textbf{1.000} & \cellcolor{green!12}\textbf{0.969} & \cellcolor{green!12}0.015 & \cellcolor{green!12}0.038 \\
 & Validation Loss & \textbf{0.700} & \textbf{1.000} & \textbf{1.000} & 0.964 & \textbf{0.013} & \textbf{0.024} \\
 & Validation MSE & 0.000 & 0.000 & 0.000 & -0.387 & 0.175 & 0.587 \\
 & Off-Manifold Norm & 0.500 & 0.800 & 0.900 & 0.922 & 0.024 & 0.050 \\
\midrule
\multirow{4}{*}{Rollouts} & \cellcolor{green!12}\method & \cellcolor{green!12}\textbf{0.200} & \cellcolor{green!12}\textbf{0.600} & \cellcolor{green!12}\textbf{0.900} & \cellcolor{green!12}\textbf{0.841} & \cellcolor{green!12}\textbf{0.044} & \cellcolor{green!12}\textbf{0.159} \\
 & Validation Loss & 0.100 & 0.100 & 0.100 & 0.244 & 0.138 & 0.291 \\
 & Validation MSE & \textbf{0.200} & 0.400 & 0.500 & 0.259 & 0.137 & 0.355 \\
 & Off-Manifold Norm & 0.100 & 0.300 & 0.600 & 0.710 & 0.069 & 0.169 \\
\midrule
\multirow{4}{*}{Noised Rollouts} & \cellcolor{green!12}\method & \cellcolor{green!12}\textbf{0.300} & \cellcolor{green!12}\textbf{0.400} & \cellcolor{green!12}\textbf{1.000} & \cellcolor{green!12}\textbf{0.857} & \cellcolor{green!12}\textbf{0.051} & \cellcolor{green!12}\textbf{0.076} \\
 & Validation Loss & 0.000 & 0.100 & 0.200 & -0.394 & 0.218 & 0.525 \\
 & Validation MSE & 0.100 & 0.300 & 0.400 & 0.148 & 0.156 & 0.434 \\
 & Off-Manifold Norm & 0.100 & 0.300 & 0.500 & 0.590 & 0.090 & 0.135 \\
\bottomrule
\end{tabular*}
\end{table}

\begin{table}[!htbp]
\centering
\small
\setlength{\tabcolsep}{3pt}
\caption{\textbf{Pouring.} Per-task ranking performance on DexMimicGen~\citep{jiang2025dexmimicgen}, averaged over $10$ seeds. Arrows indicate better values; bold marks the best reported value per metric within each validation source.}
\label{tab:per_task_pouring}
\begin{tabular*}{\linewidth}{@{\extracolsep{\fill}}llcccccc}
\toprule
Validation source & Criterion & Hit@1$\uparrow$ & Hit@3$\uparrow$ & Hit@5$\uparrow$ & $\rho\uparrow$ & MMRV$\downarrow$ & Regret$\downarrow$ \\
\midrule
\multirow{4}{*}{Held-out Demos} & \cellcolor{green!12}\method & \cellcolor{green!12}0.000 & \cellcolor{green!12}0.600 & \cellcolor{green!12}0.900 & \cellcolor{green!12}\textbf{0.797} & \cellcolor{green!12}0.013 & \cellcolor{green!12}0.123 \\
 & Validation Loss & 0.100 & 0.600 & 0.900 & 0.766 & \textbf{0.009} & 0.067 \\
 & Validation MSE & 0.000 & 0.100 & 0.400 & 0.448 & 0.021 & 0.165 \\
 & Off-Manifold Norm & \textbf{0.300} & \textbf{0.700} & \textbf{1.000} & 0.778 & \textbf{0.009} & \textbf{0.051} \\
\midrule
\multirow{4}{*}{Rollouts} & \cellcolor{green!12}\method & \cellcolor{green!12}\textbf{0.300} & \cellcolor{green!12}\textbf{0.800} & \cellcolor{green!12}\textbf{1.000} & \cellcolor{green!12}\textbf{0.803} & \cellcolor{green!12}\textbf{0.010} & \cellcolor{green!12}\textbf{0.062} \\
 & Validation Loss & 0.000 & 0.100 & 0.200 & -0.286 & 0.075 & 0.468 \\
 & Validation MSE & 0.100 & 0.100 & 0.200 & -0.009 & 0.035 & 0.339 \\
 & Off-Manifold Norm & 0.200 & 0.400 & 0.500 & 0.376 & 0.037 & 0.187 \\
\midrule
\multirow{4}{*}{Noised Rollouts} & \cellcolor{green!12}\method & \cellcolor{green!12}\textbf{0.200} & \cellcolor{green!12}\textbf{0.800} & \cellcolor{green!12}\textbf{1.000} & \cellcolor{green!12}0.609 & \cellcolor{green!12}0.025 & \cellcolor{green!12}0.137 \\
 & Validation Loss & 0.000 & 0.000 & 0.000 & -0.641 & 0.083 & 0.605 \\
 & Validation MSE & 0.000 & 0.300 & 0.800 & \textbf{0.689} & \textbf{0.015} & \textbf{0.088} \\
 & Off-Manifold Norm & 0.100 & 0.400 & 0.500 & 0.134 & 0.046 & 0.177 \\
\bottomrule
\end{tabular*}
\end{table}
\clearpage

\begin{table}[!htbp]
\centering
\small
\setlength{\tabcolsep}{3pt}
\caption{\textbf{Threading.} Per-task ranking performance on DexMimicGen~\citep{jiang2025dexmimicgen}, averaged over $10$ seeds. Arrows indicate better values; bold marks the best reported value per metric within each validation source.}
\label{tab:per_task_threading}
\begin{tabular*}{\linewidth}{@{\extracolsep{\fill}}llcccccc}
\toprule
Validation source & Criterion & Hit@1$\uparrow$ & Hit@3$\uparrow$ & Hit@5$\uparrow$ & $\rho\uparrow$ & MMRV$\downarrow$ & Regret$\downarrow$ \\
\midrule
\multirow{4}{*}{Held-out Demos} & \cellcolor{green!12}\method & \cellcolor{green!12}\textbf{0.900} & \cellcolor{green!12}\textbf{0.900} & \cellcolor{green!12}\textbf{0.900} & \cellcolor{green!12}\textbf{0.906} & \cellcolor{green!12}\textbf{0.012} & \cellcolor{green!12}\textbf{0.050} \\
 & Validation Loss & 0.300 & 0.500 & 0.700 & 0.843 & 0.028 & 0.055 \\
 & Validation MSE & 0.000 & 0.000 & 0.000 & -0.567 & 0.131 & 0.641 \\
 & Off-Manifold Norm & 0.300 & 0.400 & 0.700 & 0.778 & 0.037 & 0.138 \\
\midrule
\multirow{4}{*}{Rollouts} & \cellcolor{green!12}\method & \cellcolor{green!12}0.100 & \cellcolor{green!12}0.300 & \cellcolor{green!12}\textbf{0.800} & \cellcolor{green!12}0.591 & \cellcolor{green!12}0.055 & \cellcolor{green!12}0.239 \\
 & Validation Loss & 0.100 & 0.400 & 0.400 & 0.471 & 0.068 & 0.279 \\
 & Validation MSE & 0.000 & 0.000 & 0.100 & -0.294 & 0.110 & 0.545 \\
 & Off-Manifold Norm & \textbf{0.200} & \textbf{0.500} & 0.600 & \textbf{0.681} & \textbf{0.047} & \textbf{0.150} \\
\midrule
\multirow{4}{*}{Noised Rollouts} & \cellcolor{green!12}\method & \cellcolor{green!12}\textbf{0.400} & \cellcolor{green!12}\textbf{1.000} & \cellcolor{green!12}\textbf{1.000} & \cellcolor{green!12}\textbf{0.931} & \cellcolor{green!12}\textbf{0.016} & \cellcolor{green!12}\textbf{0.047} \\
 & Validation Loss & 0.000 & 0.200 & 0.200 & 0.543 & 0.058 & 0.257 \\
 & Validation MSE & 0.000 & 0.000 & 0.000 & 0.097 & 0.092 & 0.418 \\
 & Off-Manifold Norm & 0.100 & 0.500 & 0.600 & 0.745 & 0.036 & 0.216 \\
\bottomrule
\end{tabular*}
\end{table}

\begin{table}[!htbp]
\centering
\small
\setlength{\tabcolsep}{3pt}
\caption{\textbf{Three Piece Assembly.} Per-task ranking performance on DexMimicGen~\citep{jiang2025dexmimicgen}, averaged over $10$ seeds. Arrows indicate better values; bold marks the best reported value per metric within each validation source.}
\label{tab:per_task_three_piece_assembly}
\begin{tabular*}{\linewidth}{@{\extracolsep{\fill}}llcccccc}
\toprule
Validation source & Criterion & Hit@1$\uparrow$ & Hit@3$\uparrow$ & Hit@5$\uparrow$ & $\rho\uparrow$ & MMRV$\downarrow$ & Regret$\downarrow$ \\
\midrule
\multirow{4}{*}{Held-out Demos} & \cellcolor{green!12}\method & \cellcolor{green!12}\textbf{0.500} & \cellcolor{green!12}\textbf{0.800} & \cellcolor{green!12}\textbf{1.000} & \cellcolor{green!12}0.931 & \cellcolor{green!12}0.013 & \cellcolor{green!12}\textbf{0.045} \\
 & Validation Loss & 0.300 & \textbf{0.800} & \textbf{1.000} & \textbf{0.953} & \textbf{0.009} & 0.055 \\
 & Validation MSE & 0.000 & 0.100 & 0.200 & 0.368 & 0.067 & 0.302 \\
 & Off-Manifold Norm & 0.100 & 0.600 & \textbf{1.000} & 0.877 & 0.024 & 0.079 \\
\midrule
\multirow{4}{*}{Rollouts} & \cellcolor{green!12}\method & \cellcolor{green!12}0.100 & \cellcolor{green!12}\textbf{0.300} & \cellcolor{green!12}\textbf{0.700} & \cellcolor{green!12}\textbf{0.703} & \cellcolor{green!12}\textbf{0.041} & \cellcolor{green!12}\textbf{0.163} \\
 & Validation Loss & 0.000 & 0.100 & 0.100 & 0.218 & 0.090 & 0.366 \\
 & Validation MSE & 0.000 & 0.000 & 0.000 & -0.378 & 0.130 & 0.511 \\
 & Off-Manifold Norm & \textbf{0.200} & \textbf{0.300} & 0.600 & 0.671 & 0.048 & 0.177 \\
\midrule
\multirow{4}{*}{Noised Rollouts} & \cellcolor{green!12}\method & \cellcolor{green!12}\textbf{0.300} & \cellcolor{green!12}\textbf{0.600} & \cellcolor{green!12}\textbf{0.800} & \cellcolor{green!12}\textbf{0.869} & \cellcolor{green!12}\textbf{0.024} & \cellcolor{green!12}\textbf{0.081} \\
 & Validation Loss & 0.000 & 0.000 & 0.100 & -0.427 & 0.148 & 0.575 \\
 & Validation MSE & 0.000 & 0.100 & 0.100 & -0.023 & 0.104 & 0.365 \\
 & Off-Manifold Norm & 0.100 & 0.300 & 0.300 & 0.479 & 0.074 & 0.305 \\
\bottomrule
\end{tabular*}
\end{table}
\clearpage

\clearpage
\subsection{Real-World Experiments}
\label{app:real_per_task}
\begin{figure}[!htbp]
\centering
\includegraphics[width=0.49\linewidth]{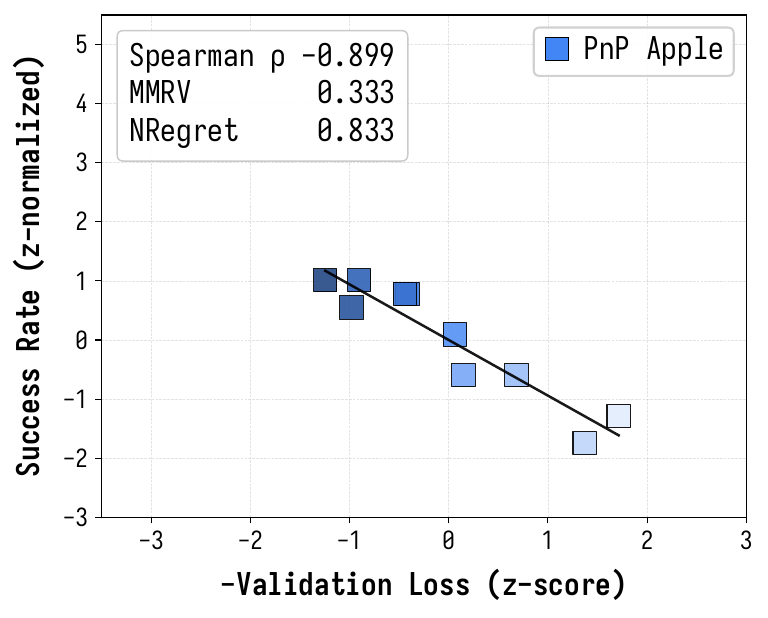}\hfill
\includegraphics[width=0.49\linewidth]{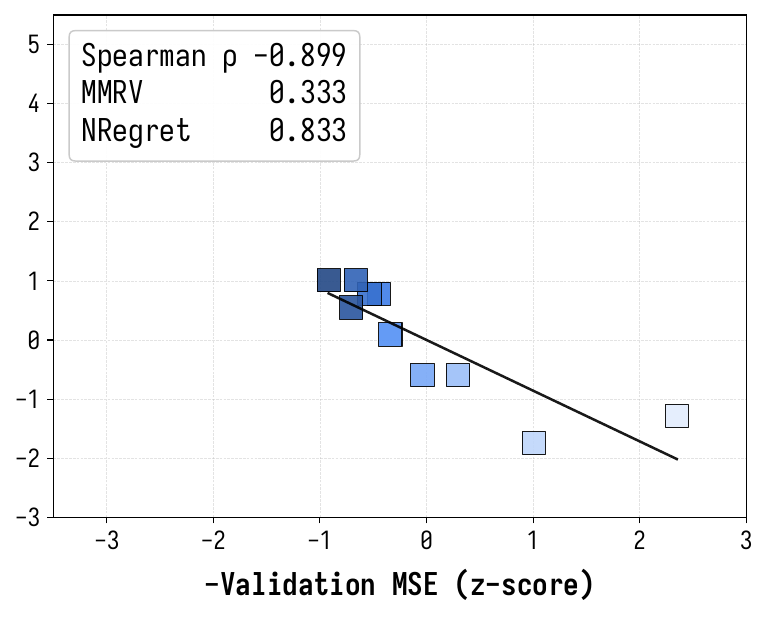}
\par\smallskip
\includegraphics[width=0.49\linewidth]{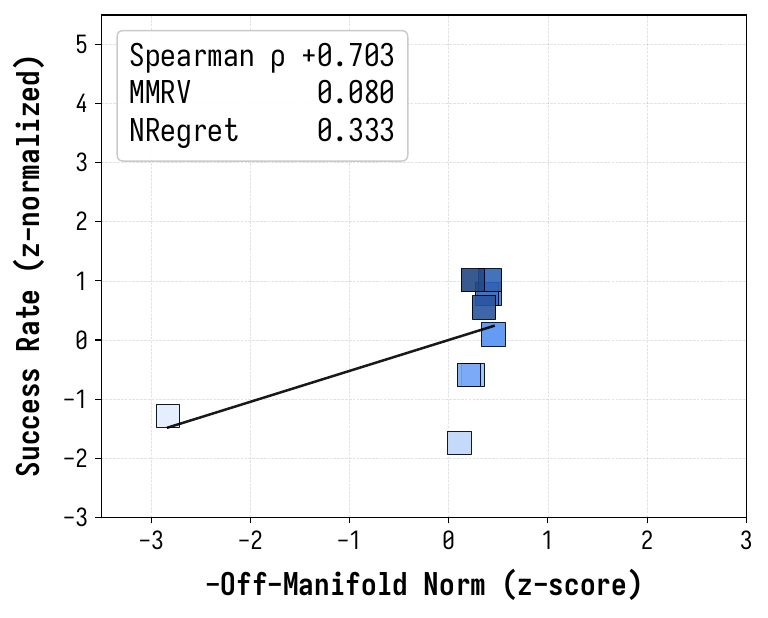}\hfill
\includegraphics[width=0.49\linewidth]{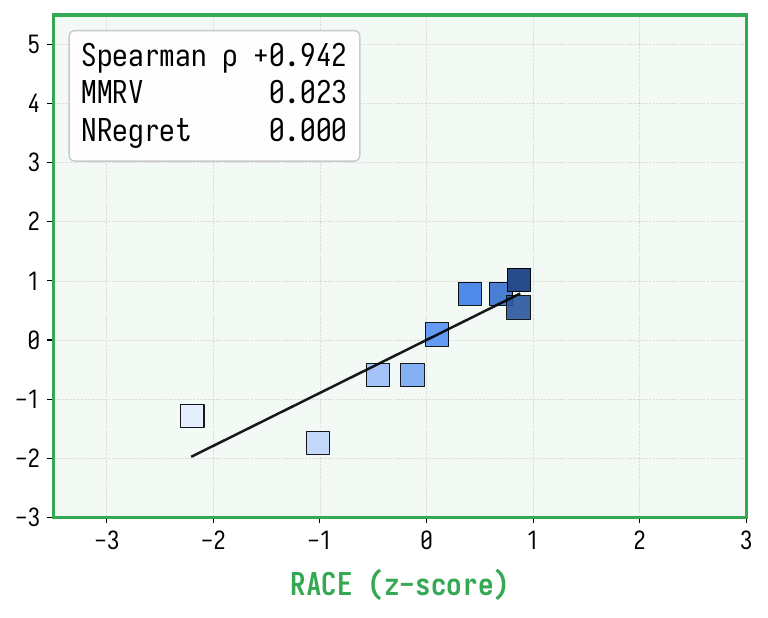}
\caption{\textbf{Diffusion Policy: PnP Apple.} Task-level checkpoint-score versus success-rate scatter plots. Panels show Validation Loss, Validation MSE, Off-Manifold Norm, and \method from left to right and top to bottom; both axes use within-task z-scores.}
\label{fig:dp-pnp-apple}
\end{figure}

\begin{figure}[!htbp]
\centering
\includegraphics[width=0.49\linewidth]{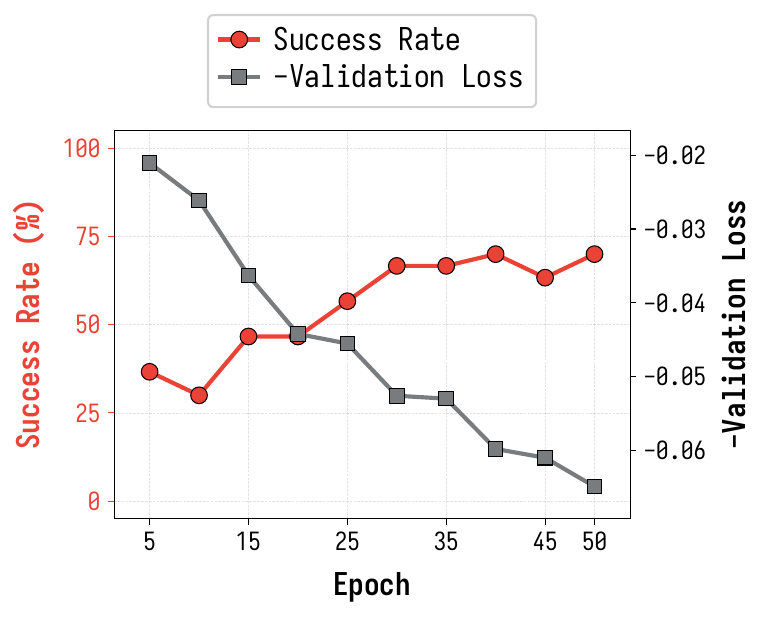}\hfill
\includegraphics[width=0.49\linewidth]{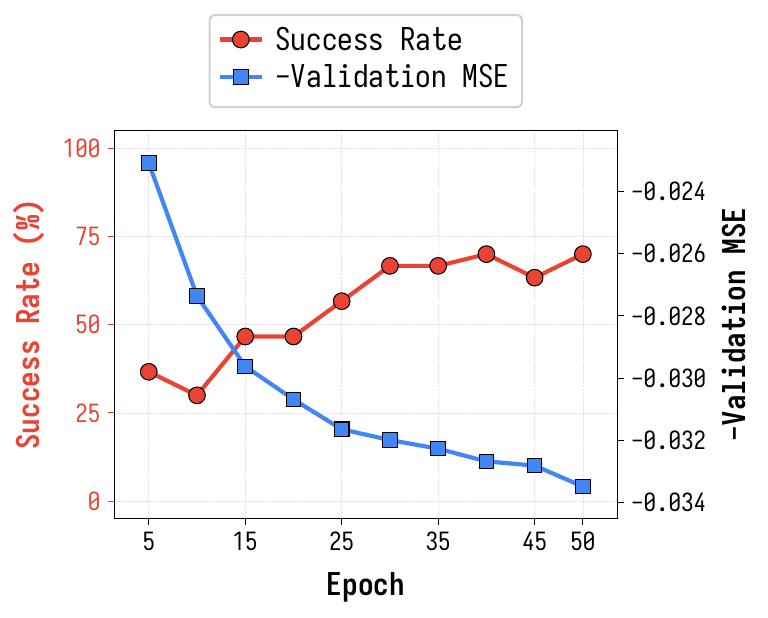}
\par\smallskip
\includegraphics[width=0.49\linewidth]{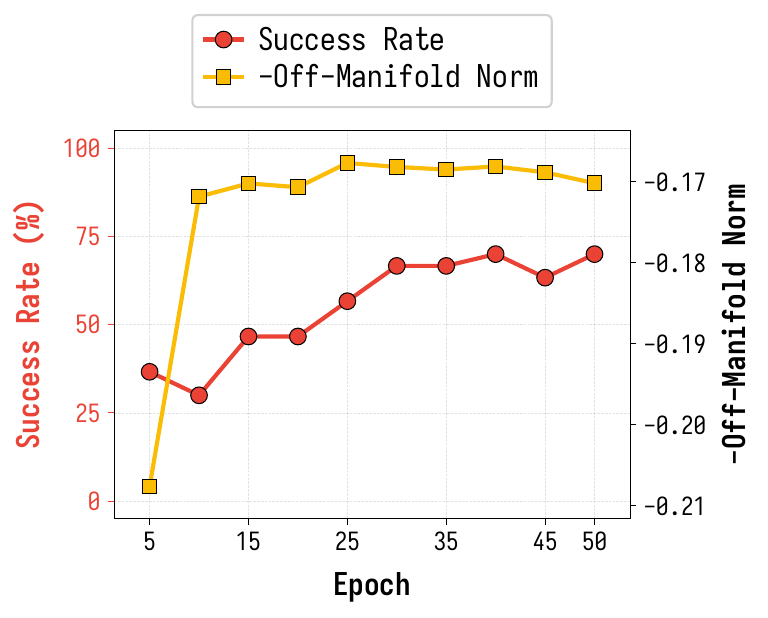}\hfill
\includegraphics[width=0.49\linewidth]{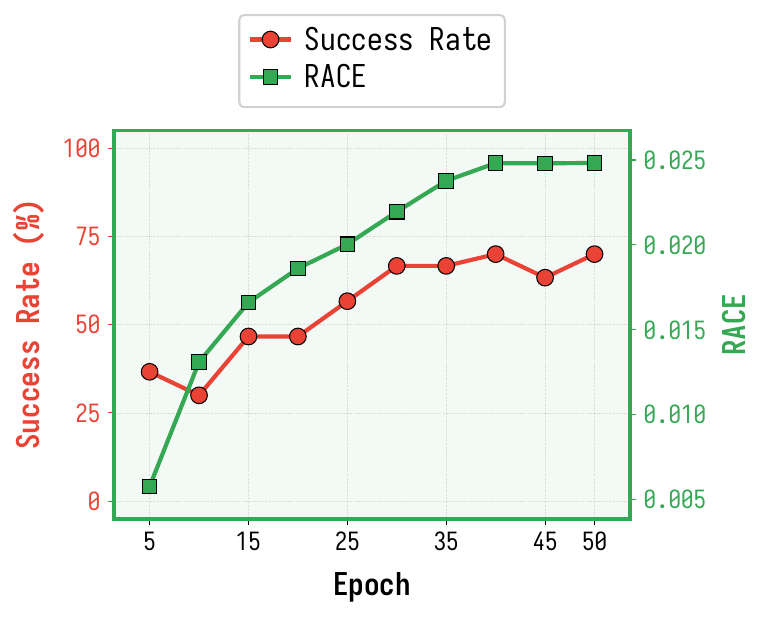}
\caption{\textbf{Diffusion Policy: PnP Apple across training.} Each panel overlays empirical success rate (left axis, percent) and a signed criterion value (right axis) against training epoch. Panel order matches the preceding task-level scatter plots, where higher criterion values are preferred.}
\label{fig:dp-pnp-apple-progress}
\end{figure}

\begin{figure}[!htbp]
\centering
\includegraphics[width=0.49\linewidth]{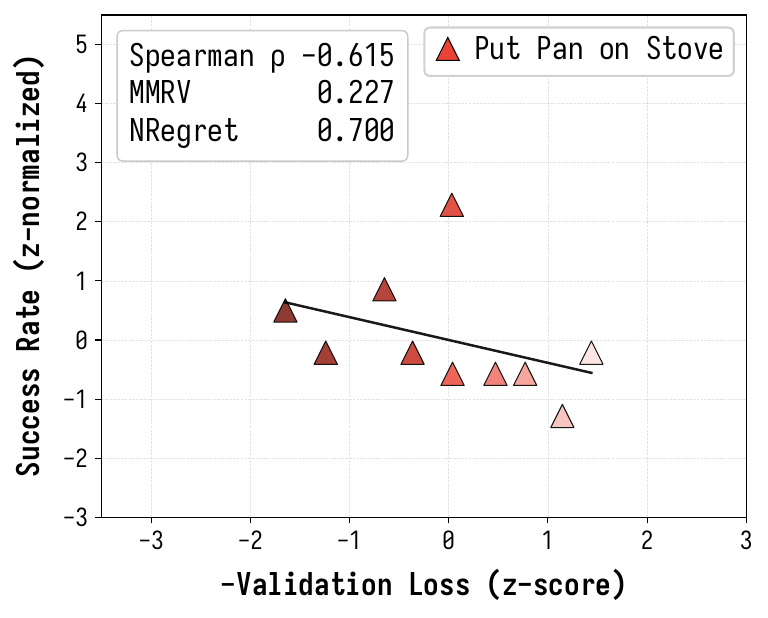}\hfill
\includegraphics[width=0.49\linewidth]{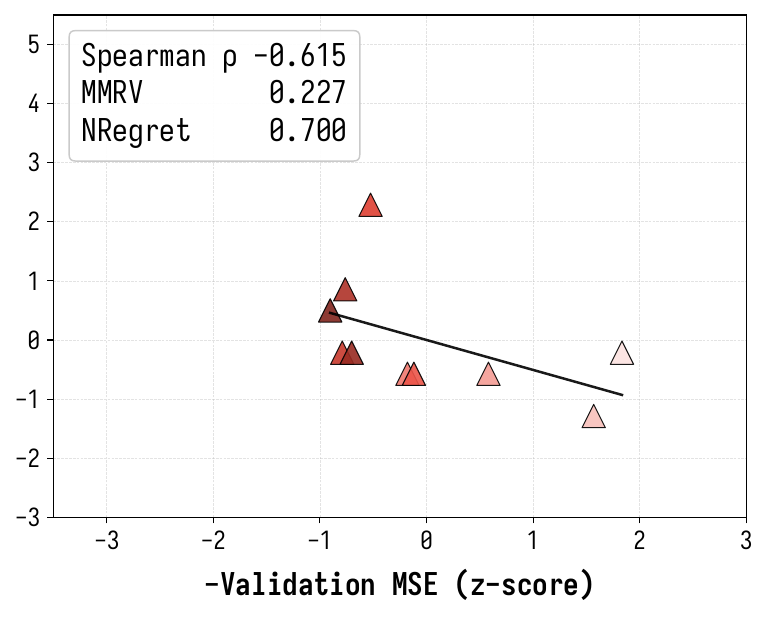}
\par\smallskip
\includegraphics[width=0.49\linewidth]{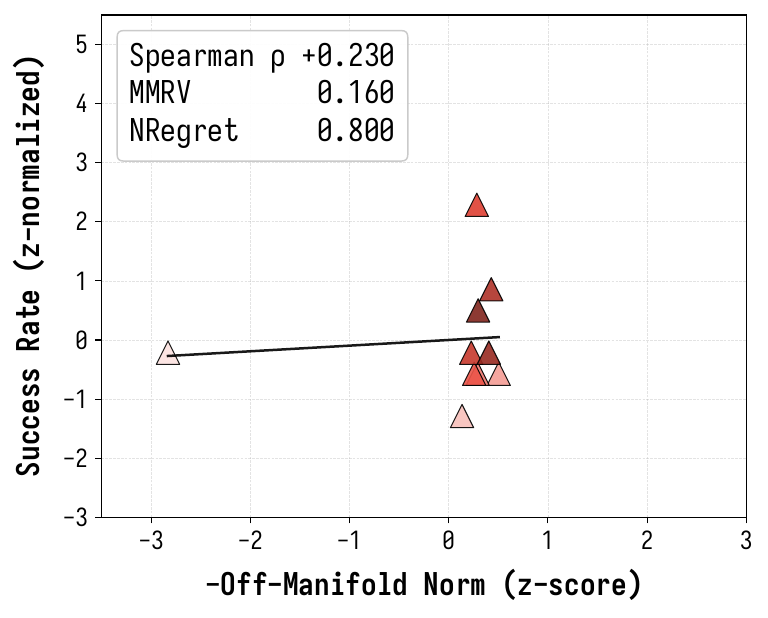}\hfill
\includegraphics[width=0.49\linewidth]{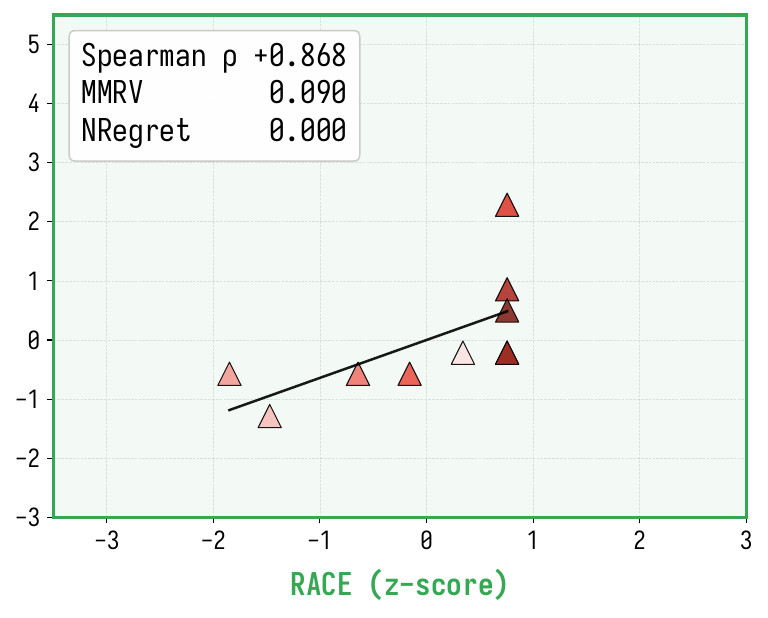}
\caption{\textbf{Diffusion Policy: Put Pan on Stove.} Task-level checkpoint-score versus success-rate scatter plots. Panels show Validation Loss, Validation MSE, Off-Manifold Norm, and \method from left to right and top to bottom; both axes use within-task z-scores.}
\label{fig:dp-put-pan}
\end{figure}

\begin{figure}[!htbp]
\centering
\includegraphics[width=0.49\linewidth]{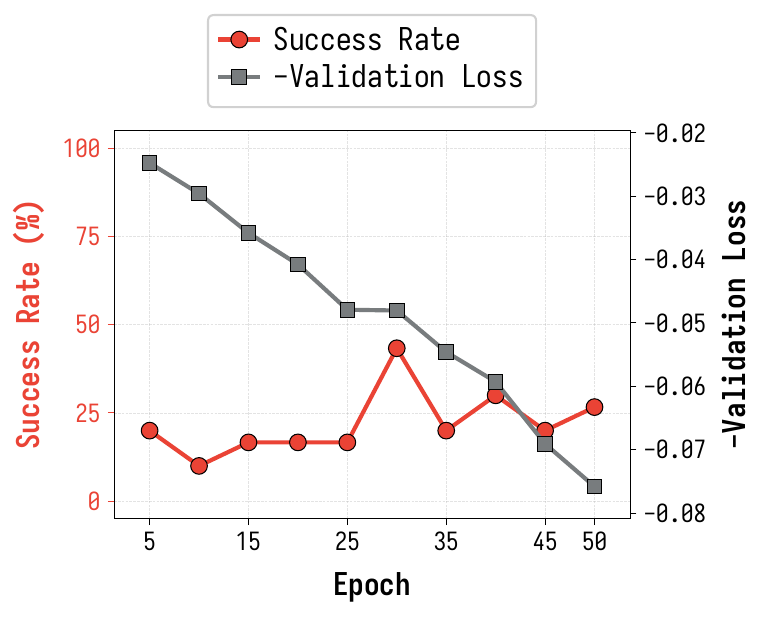}\hfill
\includegraphics[width=0.49\linewidth]{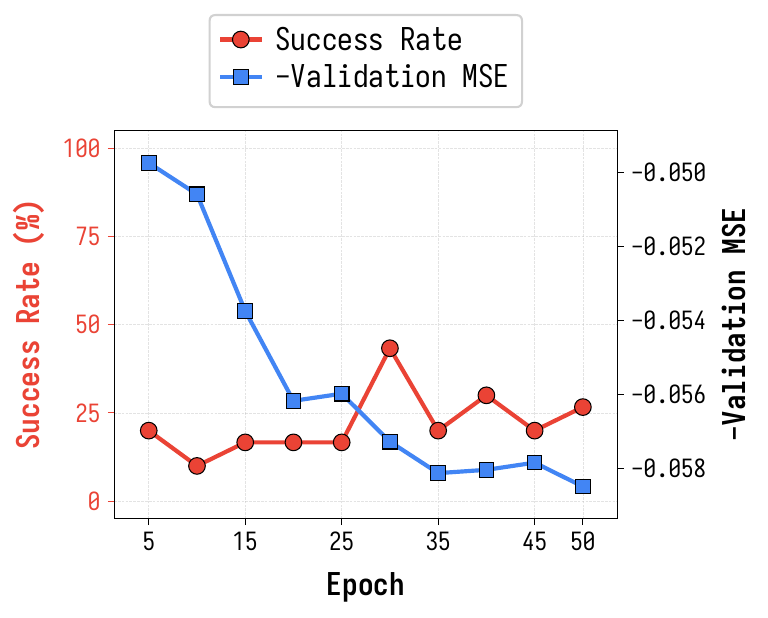}
\par\smallskip
\includegraphics[width=0.49\linewidth]{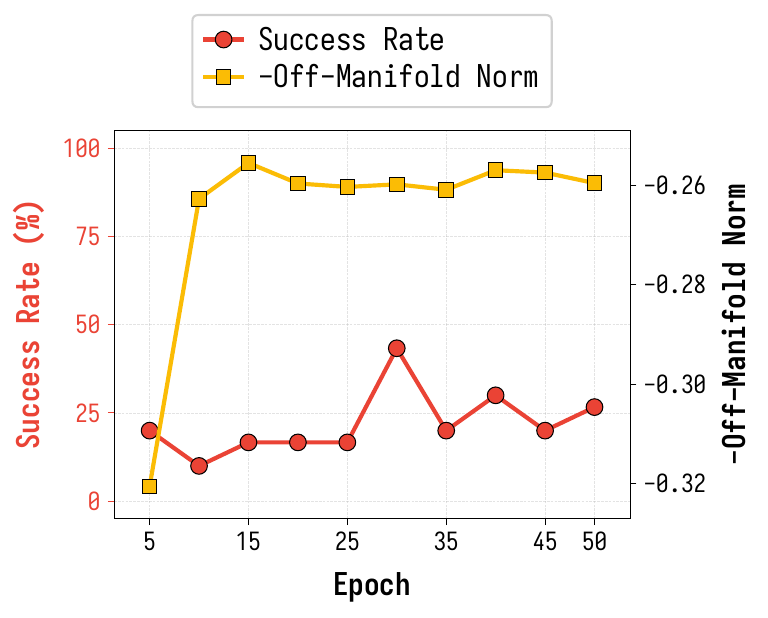}\hfill
\includegraphics[width=0.49\linewidth]{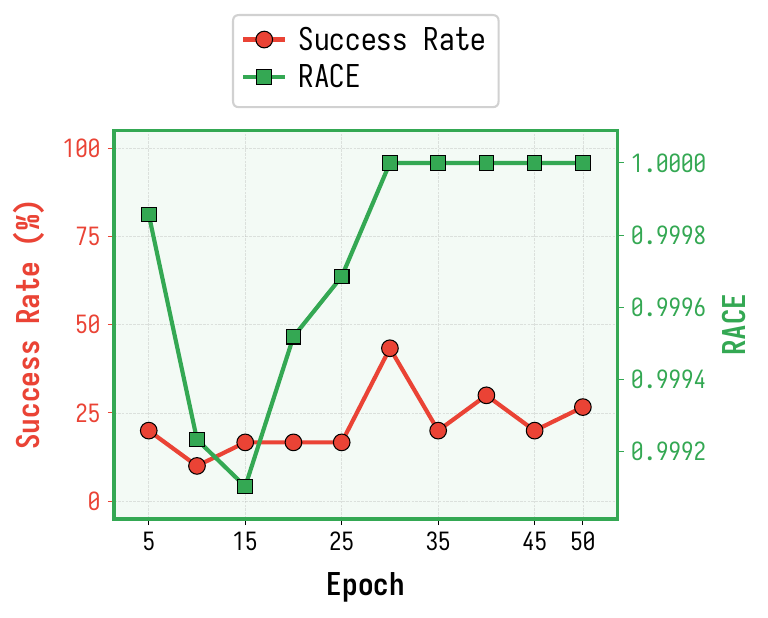}
\caption{\textbf{Diffusion Policy: Put Pan on Stove across training.} Each panel overlays empirical success rate (left axis, percent) and a signed criterion value (right axis) against training epoch. Panel order matches the preceding task-level scatter plots, where higher criterion values are preferred.}
\label{fig:dp-put-pan-progress}
\end{figure}

\begin{figure}[!htbp]
\centering
\includegraphics[width=0.49\linewidth]{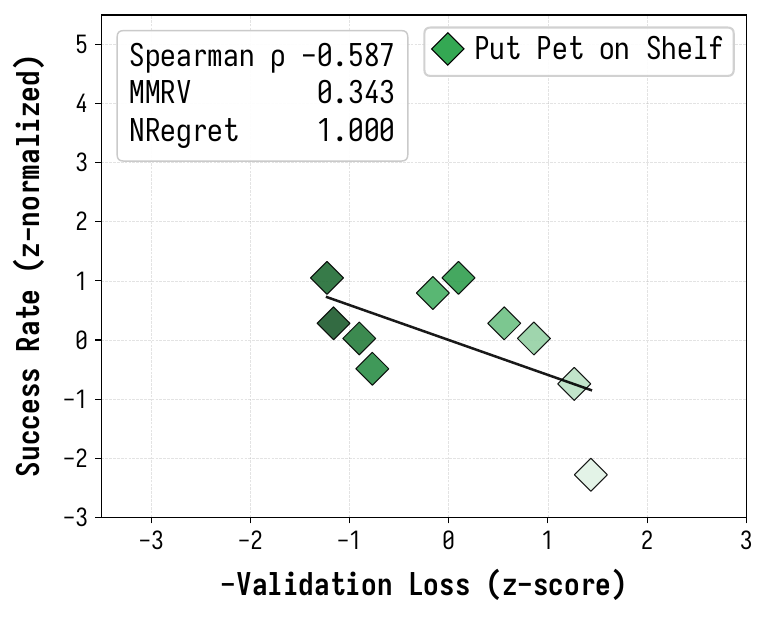}\hfill
\includegraphics[width=0.49\linewidth]{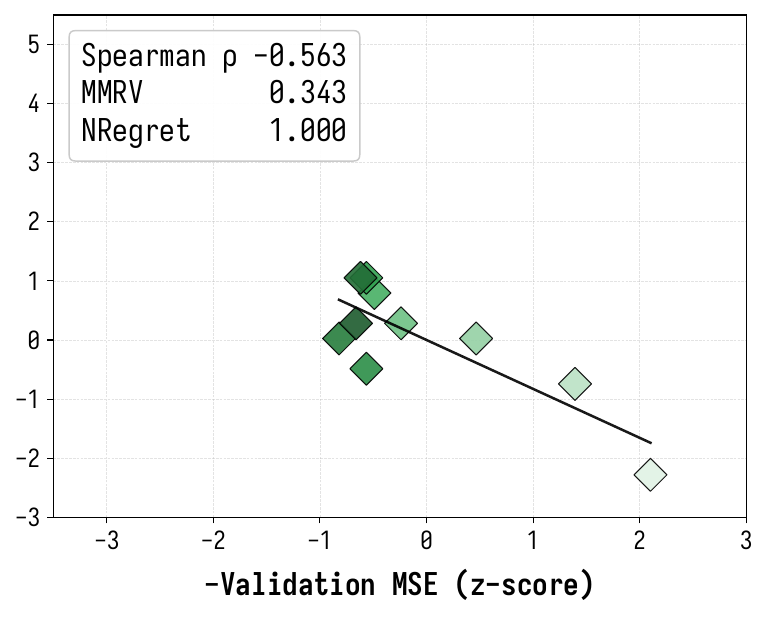}
\par\smallskip
\includegraphics[width=0.49\linewidth]{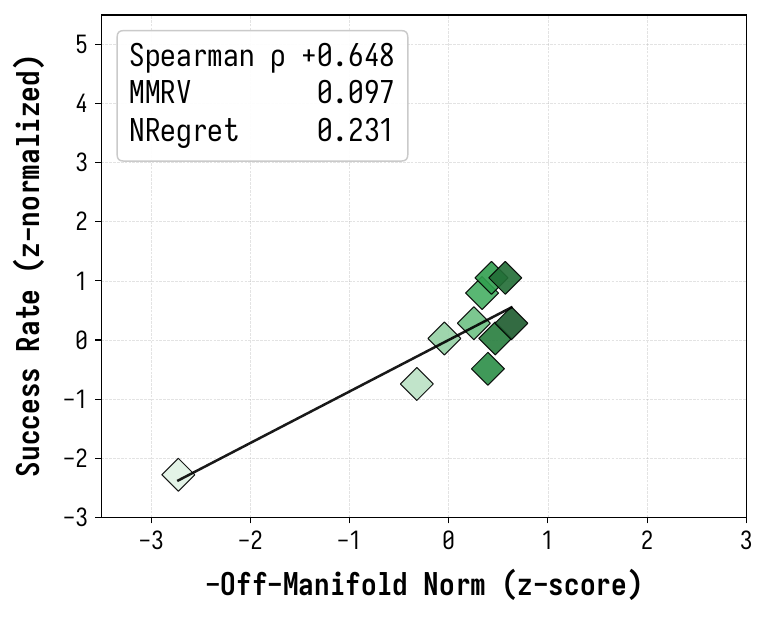}\hfill
\includegraphics[width=0.49\linewidth]{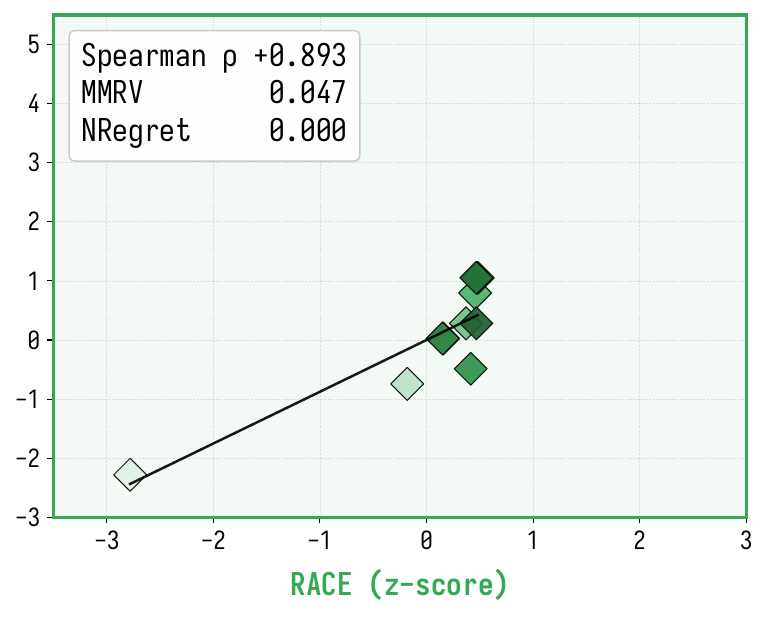}
\caption{\textbf{Diffusion Policy: Put Pet on Shelf.} Task-level checkpoint-score versus success-rate scatter plots. Panels show Validation Loss, Validation MSE, Off-Manifold Norm, and \method from left to right and top to bottom; both axes use within-task z-scores.}
\label{fig:dp-put-pet}
\end{figure}

\begin{figure}[!htbp]
\centering
\includegraphics[width=0.49\linewidth]{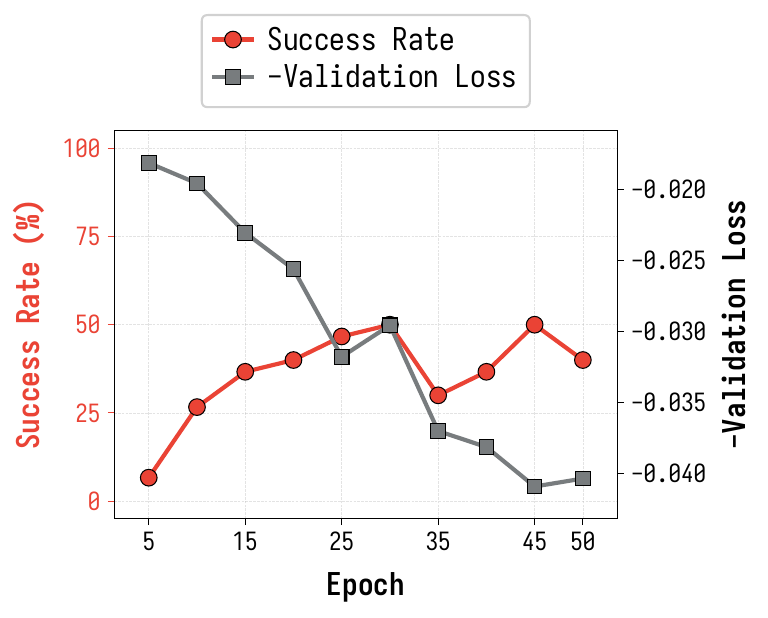}\hfill
\includegraphics[width=0.49\linewidth]{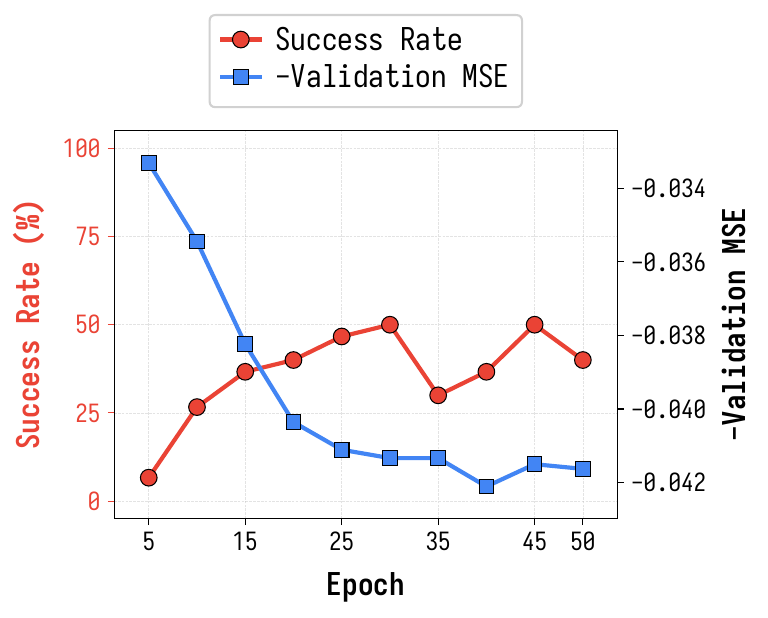}
\par\smallskip
\includegraphics[width=0.49\linewidth]{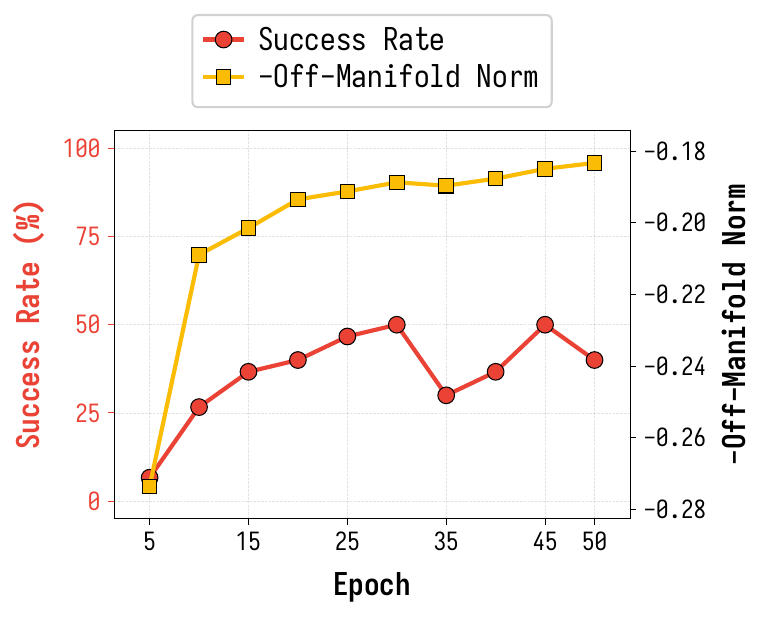}\hfill
\includegraphics[width=0.49\linewidth]{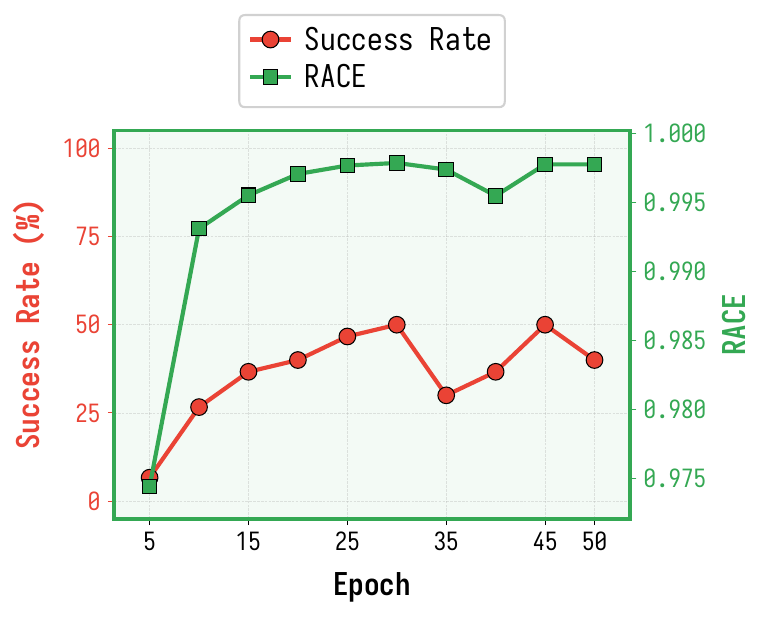}
\caption{\textbf{Diffusion Policy: Put Pet on Shelf across training.} Each panel overlays empirical success rate (left axis, percent) and a signed criterion value (right axis) against training epoch. Panel order matches the preceding task-level scatter plots, where higher criterion values are preferred.}
\label{fig:dp-put-pet-progress}
\end{figure}

\begin{figure}[!htbp]
\centering
\includegraphics[width=0.49\linewidth]{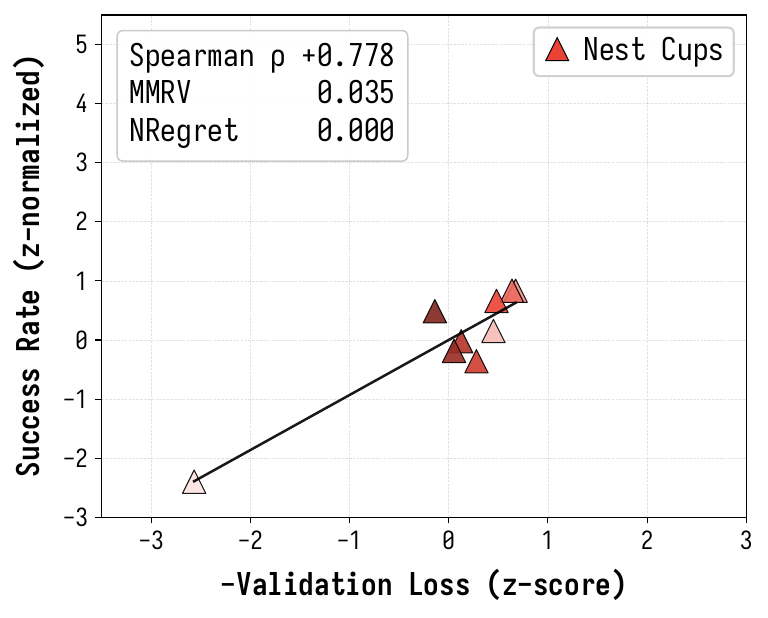}\hfill
\includegraphics[width=0.49\linewidth]{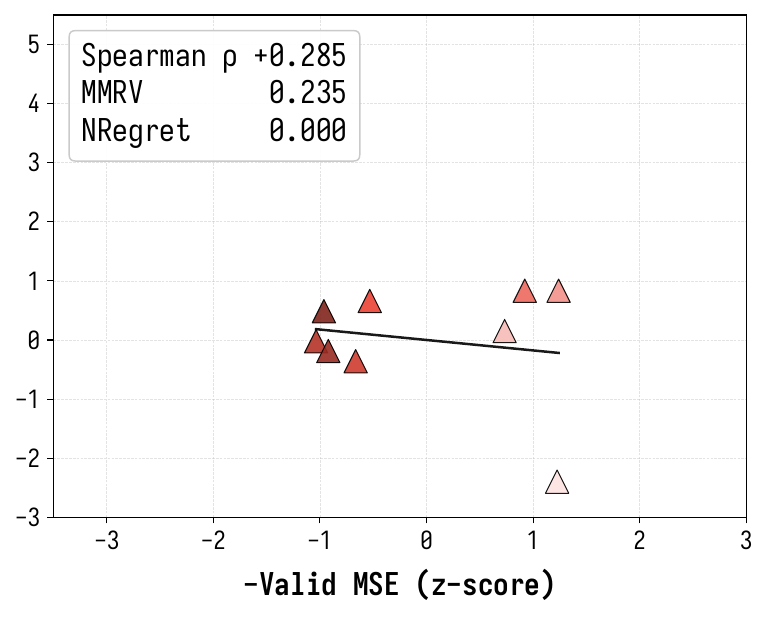}
\par\smallskip
\includegraphics[width=0.49\linewidth]{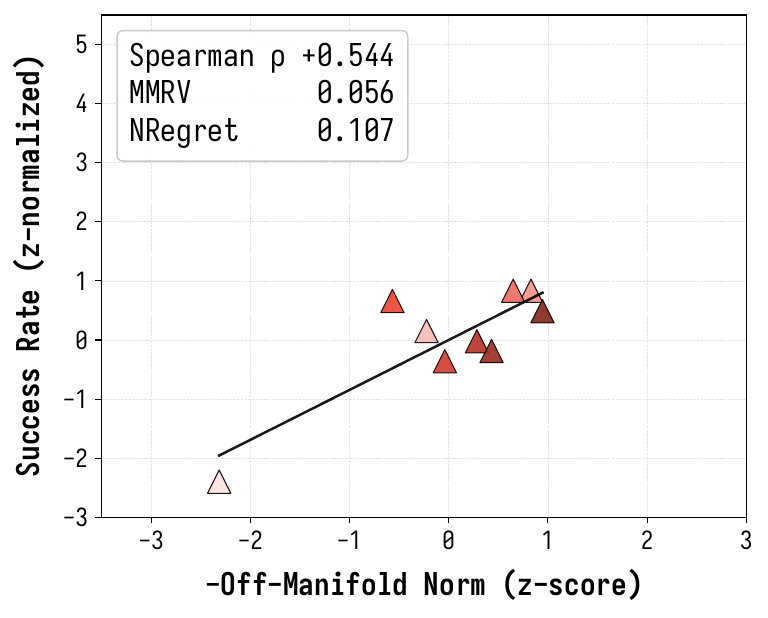}\hfill
\includegraphics[width=0.49\linewidth]{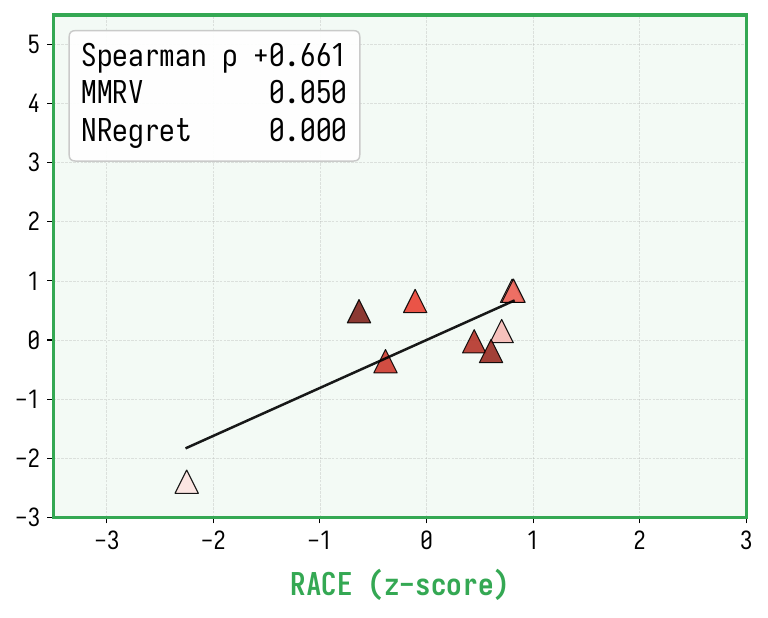}
\caption{\textbf{$\pi_{0.5}$-DROID: Nest Cups.} Task-level checkpoint-score versus success-rate scatter plots. Panels show Validation Loss, Validation MSE, Off-Manifold Norm, and \method from left to right and top to bottom; both axes use within-task z-scores.}
\label{fig:vla-nest-cups}
\end{figure}

\begin{figure}[!htbp]
\centering
\includegraphics[width=0.49\linewidth]{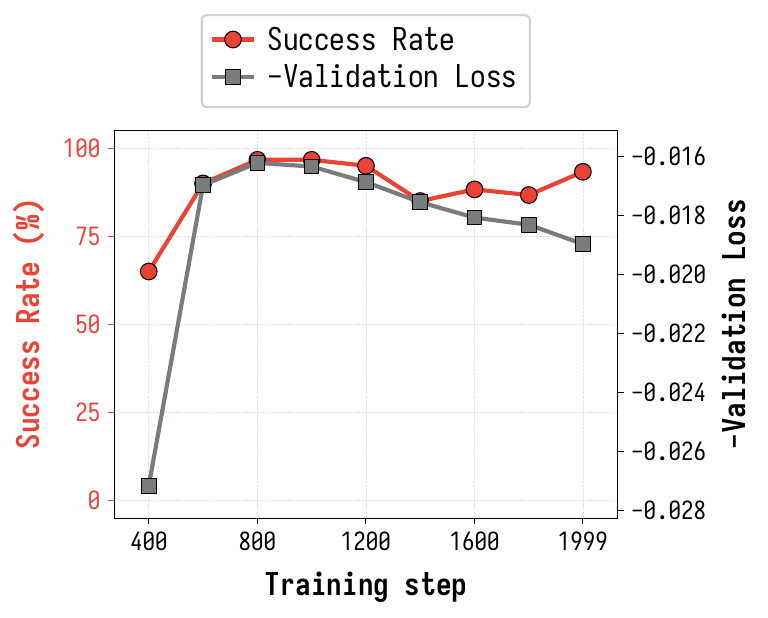}\hfill
\includegraphics[width=0.49\linewidth]{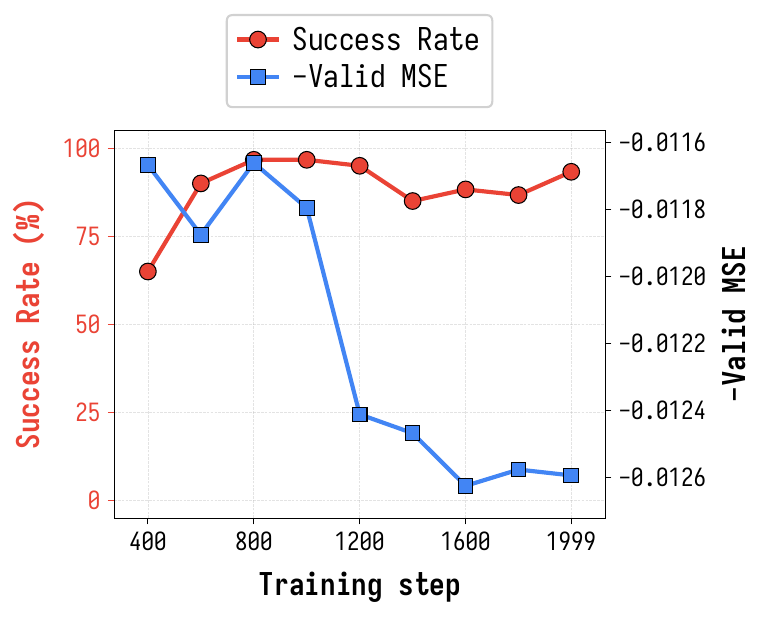}
\par\smallskip
\includegraphics[width=0.49\linewidth]{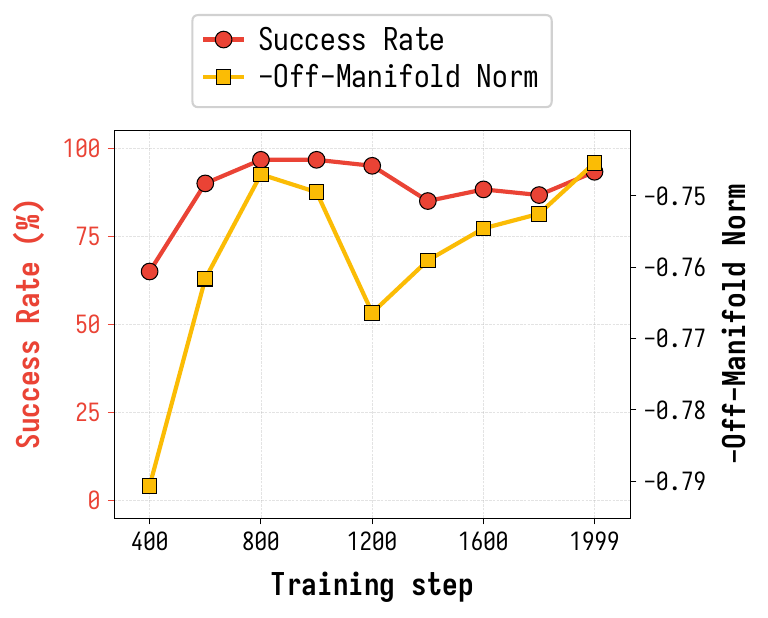}\hfill
\includegraphics[width=0.49\linewidth]{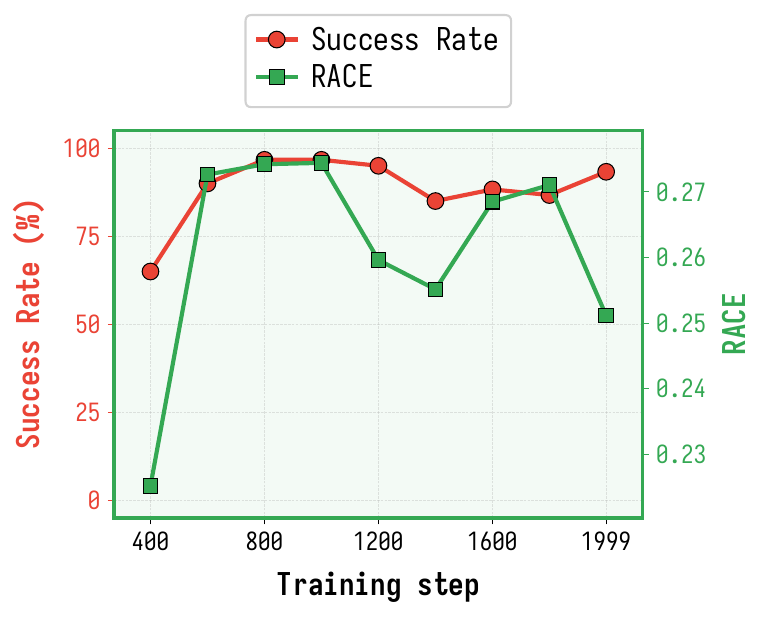}
\caption{\textbf{$\pi_{0.5}$-DROID: Nest Cups across training.} Each panel overlays empirical success rate (left axis, percent) and a signed criterion value (right axis) against training step. Panel order matches the preceding task-level scatter plots, where higher criterion values are preferred.}
\label{fig:vla-nest-cups-progress}
\end{figure}

\begin{figure}[!htbp]
\centering
\includegraphics[width=0.49\linewidth]{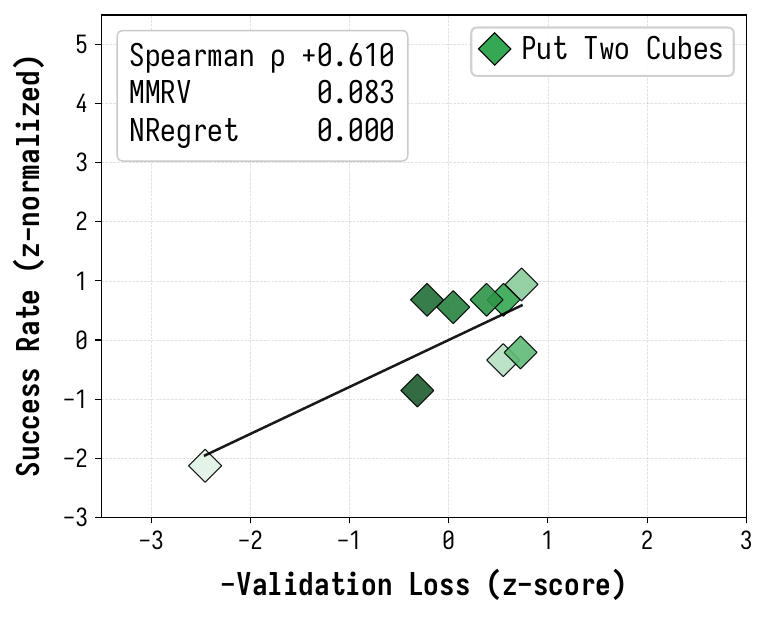}\hfill
\includegraphics[width=0.49\linewidth]{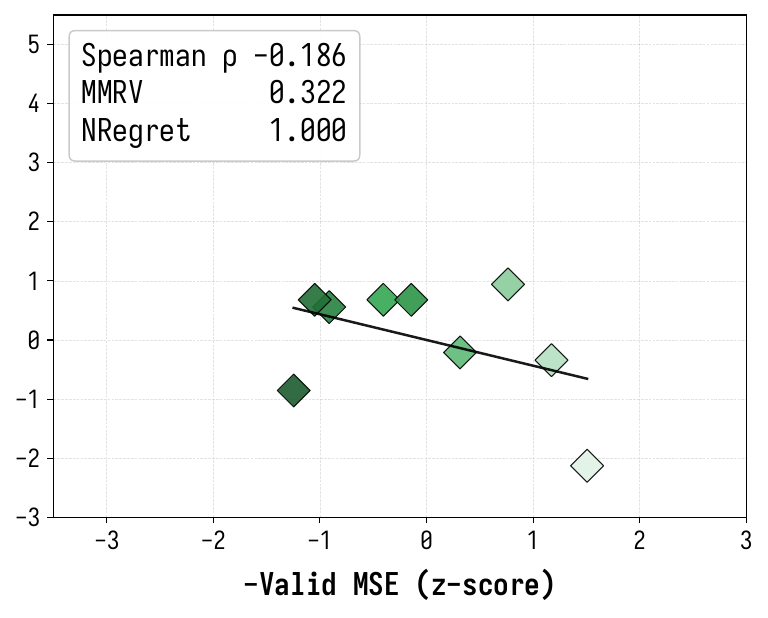}
\par\smallskip
\includegraphics[width=0.49\linewidth]{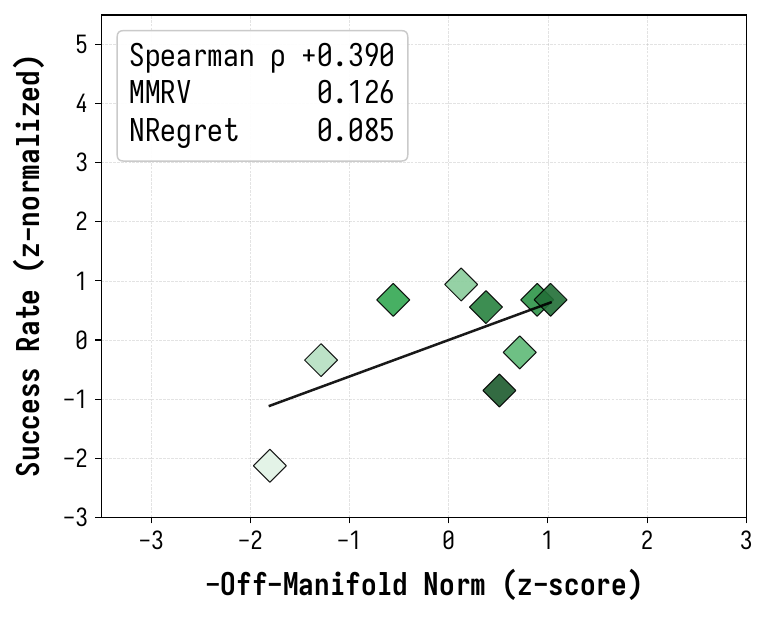}\hfill
\includegraphics[width=0.49\linewidth]{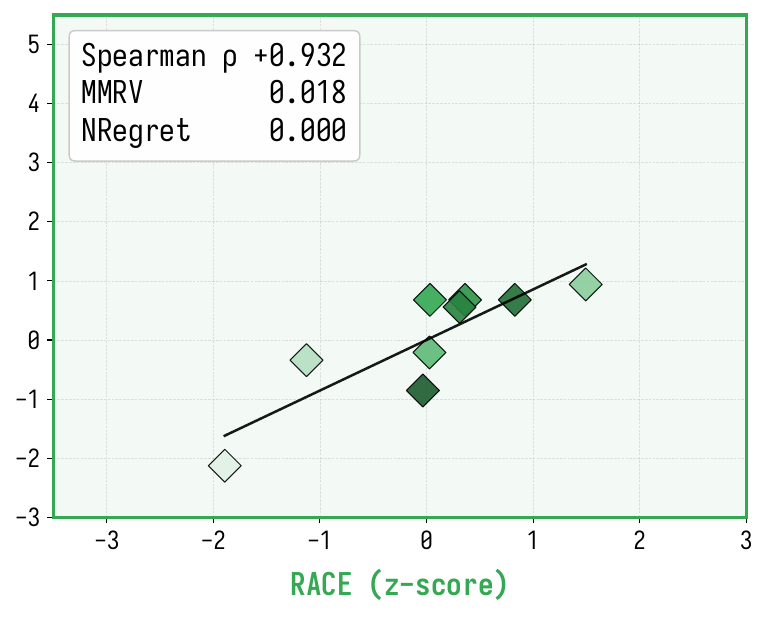}
\caption{\textbf{$\pi_{0.5}$-DROID: Put Two Cubes.} Task-level checkpoint-score versus success-rate scatter plots. Panels show Validation Loss, Validation MSE, Off-Manifold Norm, and \method from left to right and top to bottom; both axes use within-task z-scores.}
\label{fig:vla-put-two-cubes}
\end{figure}

\begin{figure}[!htbp]
\centering
\includegraphics[width=0.49\linewidth]{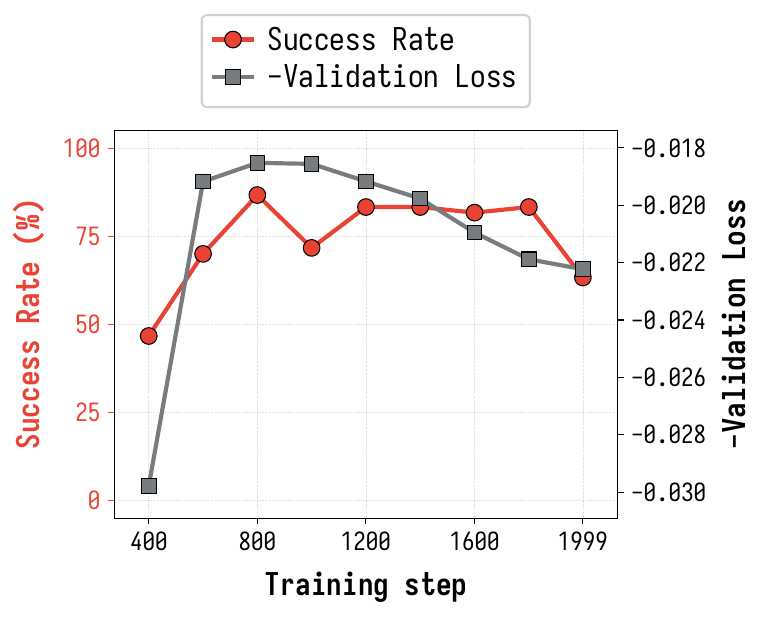}\hfill
\includegraphics[width=0.49\linewidth]{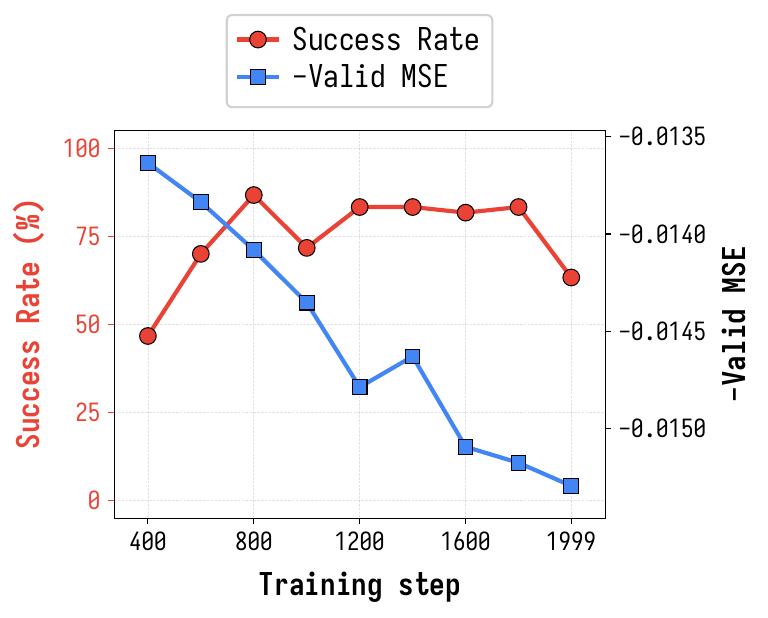}
\par\smallskip
\includegraphics[width=0.49\linewidth]{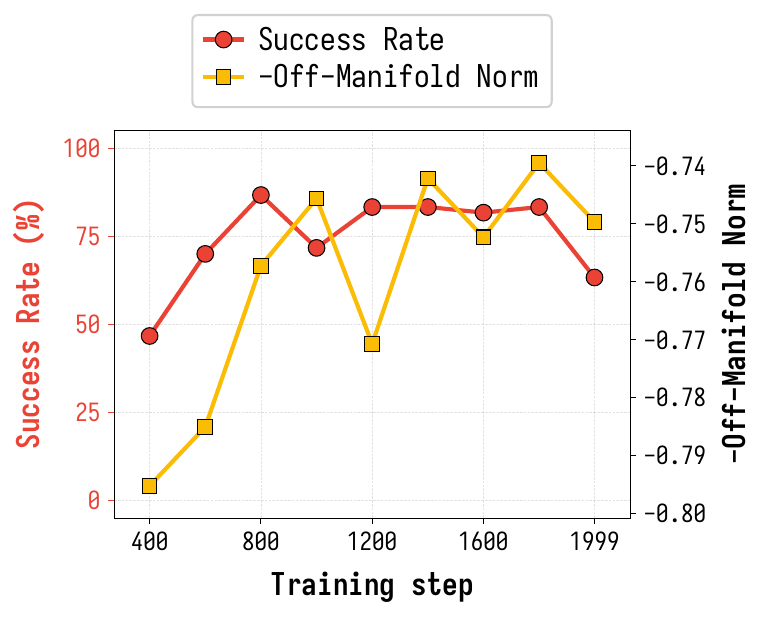}\hfill
\includegraphics[width=0.49\linewidth]{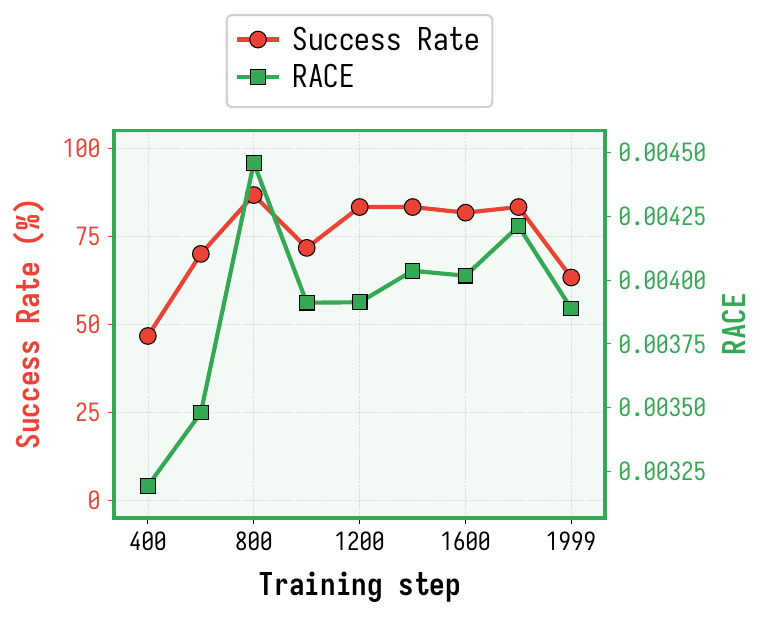}
\caption{\textbf{$\pi_{0.5}$-DROID: Put Two Cubes across training.} Each panel overlays empirical success rate (left axis, percent) and a signed criterion value (right axis) against training step. Panel order matches the preceding task-level scatter plots, where higher criterion values are preferred.}
\label{fig:vla-put-two-cubes-progress}
\end{figure}

\begin{figure}[!htbp]
\centering
\includegraphics[width=0.49\linewidth]{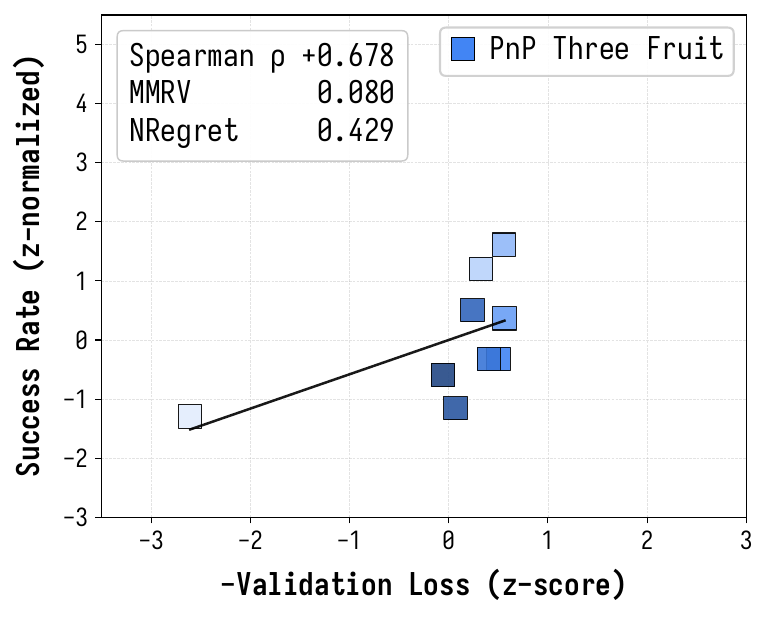}\hfill
\includegraphics[width=0.49\linewidth]{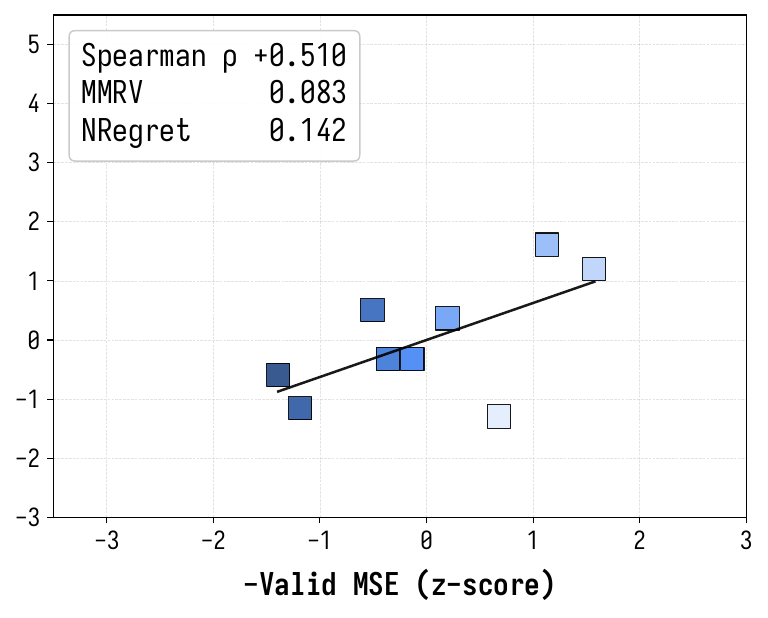}
\par\smallskip
\includegraphics[width=0.49\linewidth]{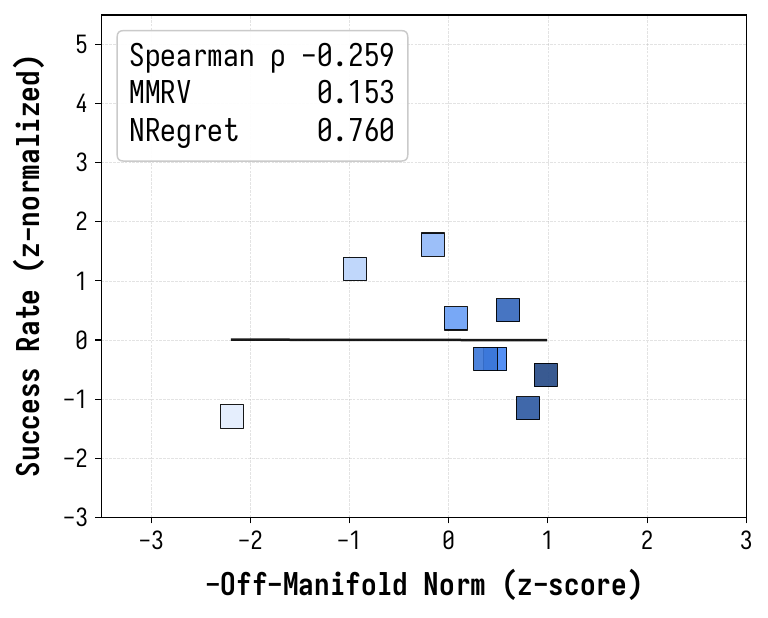}\hfill
\includegraphics[width=0.49\linewidth]{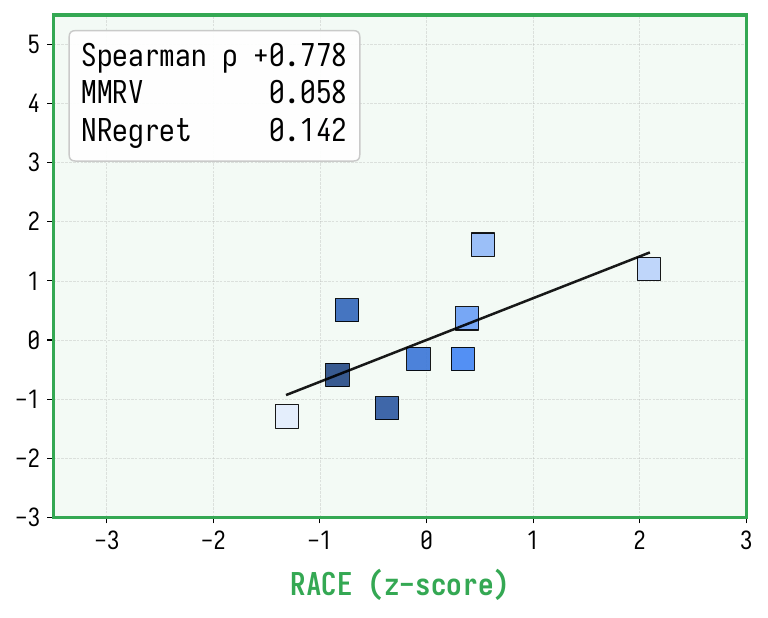}
\caption{\textbf{$\pi_{0.5}$-DROID: PnP 3 Fruit.} Task-level checkpoint-score versus success-rate scatter plots. Panels show Validation Loss, Validation MSE, Off-Manifold Norm, and \method from left to right and top to bottom; both axes use within-task z-scores.}
\label{fig:vla-pnp-three-fruit}
\end{figure}

\begin{figure}[!htbp]
\centering
\includegraphics[width=0.49\linewidth]{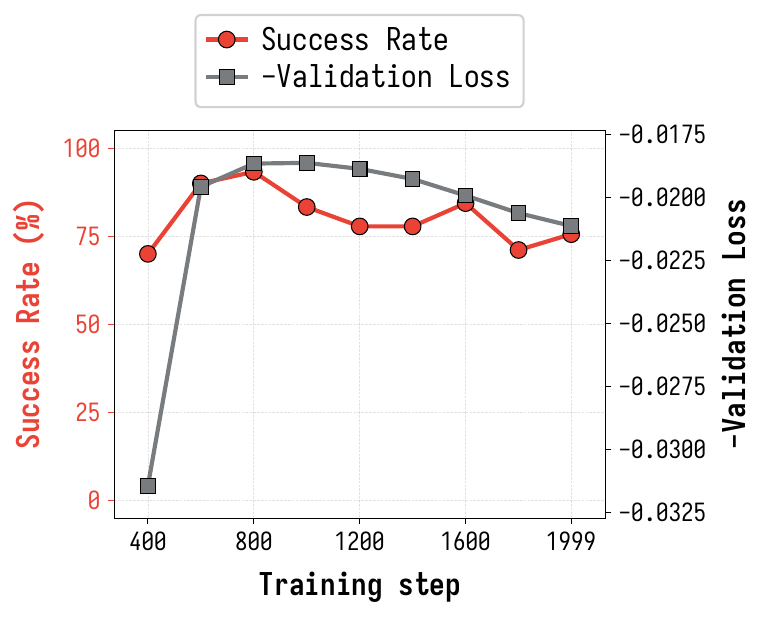}\hfill
\includegraphics[width=0.49\linewidth]{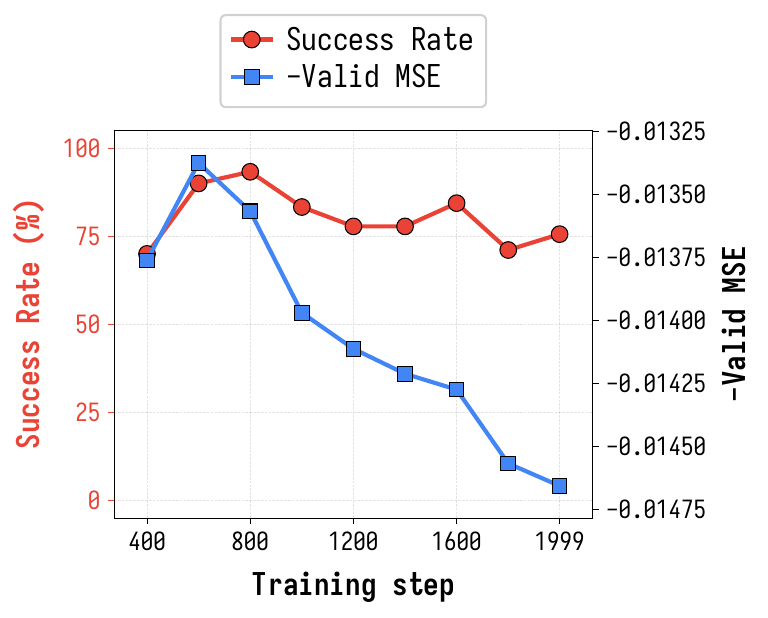}
\par\smallskip
\includegraphics[width=0.49\linewidth]{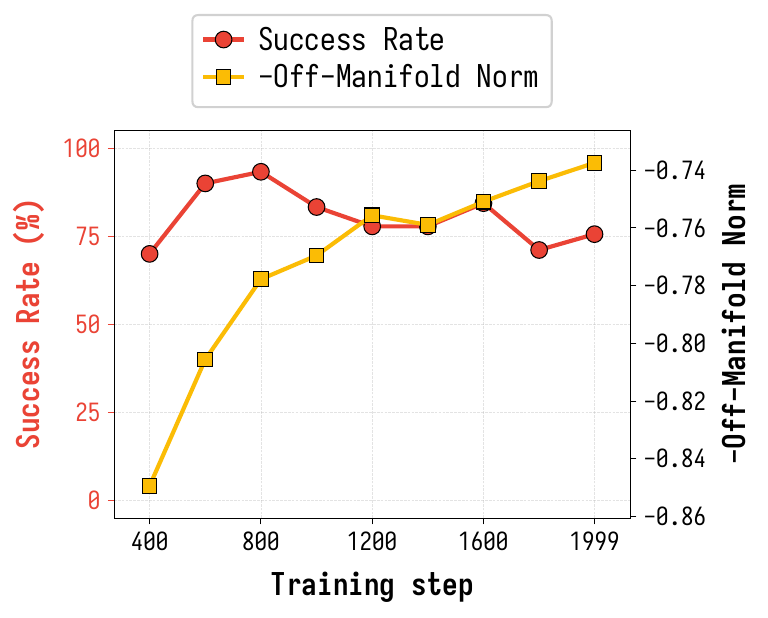}\hfill
\includegraphics[width=0.49\linewidth]{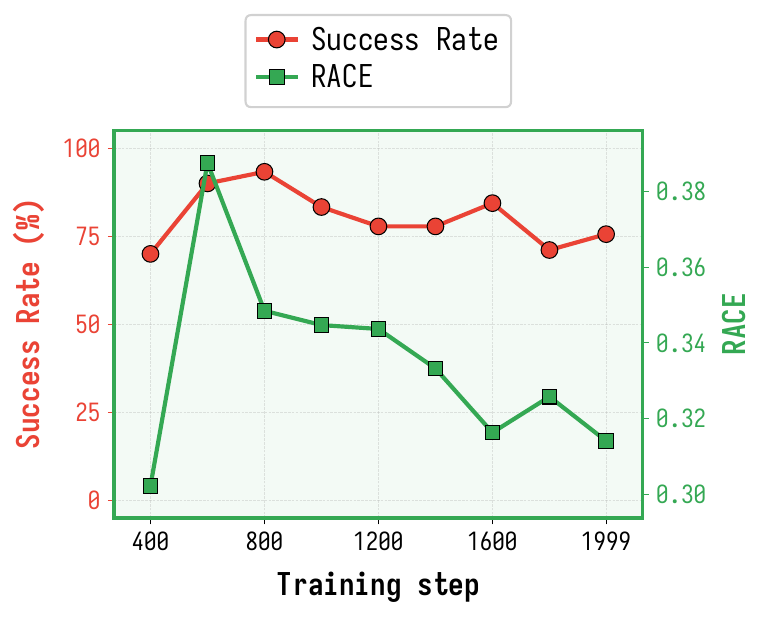}
\caption{\textbf{$\pi_{0.5}$-DROID: PnP 3 Fruit across training.} Each panel overlays empirical success rate (left axis, percent) and a signed criterion value (right axis) against training step. Panel order matches the preceding task-level scatter plots, where higher criterion values are preferred.}
\label{fig:vla-pnp-three-fruit-progress}
\end{figure}

Each task has two figures, each containing panels for the four criteria: one shows checkpoint-level scatter plots, and the other shows training progress. For Diffusion Policy, \Cref{fig:dp-pnp-apple,fig:dp-pnp-apple-progress} show \emph{PnP Apple}, \Cref{fig:dp-put-pan,fig:dp-put-pan-progress} show \emph{Put Pan on Stove}, and \Cref{fig:dp-put-pet,fig:dp-put-pet-progress} show \emph{Put Pet on Shelf}. For $\pi_{0.5}$-DROID, \Cref{fig:vla-nest-cups,fig:vla-nest-cups-progress} show \emph{Nest Cups}, \Cref{fig:vla-put-two-cubes,fig:vla-put-two-cubes-progress} show \emph{Put Two Cubes}, and \Cref{fig:vla-pnp-three-fruit,fig:vla-pnp-three-fruit-progress} show \emph{PnP 3 Fruit}.

In the scatter plots, each point is a checkpoint; success rates and criterion scores are standardized separately within each task using sample standard deviations. All baseline scores are sign-negated, so higher values are preferred for every criterion. The displayed Spearman correlation, MMRV, and nRegret are descriptive task-level point estimates, and regression lines visualize trends. In the progress plots, the left axis gives empirical success rate in percent and the right axis gives the signed, unstandardized criterion value. The horizontal axis is training epoch for Diffusion Policy and training step for $\pi_{0.5}$-DROID. These figures retain the supplied checkpoint set: ten checkpoints for each Diffusion Policy task and a displayed subset of nine checkpoints for each $\pi_{0.5}$-DROID task. The VLA curves shown here cover only this supplied subset of the checkpoint evaluations described in \Cref{sec:real_exp}.

The two views distinguish overall checkpoint-order agreement from the timing of a criterion's preferred checkpoint. A criterion can correlate with early improvement while missing a later decline or success-rate peak; the progress plots expose such differences without treating decreasing supervised loss as evidence of increasing deployment success.

\clearpage

\clearpage

\clearpage

\clearpage

\clearpage

\clearpage

\clearpage

\clearpage

\clearpage

\clearpage

\clearpage

\clearpage

\end{document}